\documentclass[sigconf]{acmart}

\usepackage[normalem]{ulem}
\useunder{\uline}{\ul}{}
\usepackage{multirow}
\usepackage{makecell}
\usepackage{pifont}
\usepackage{diagbox}
\usepackage{subfigure}
\usepackage{algorithm}
\usepackage{listings}
\usepackage{enumitem}
\setlist{nosep}

\newtheorem*{theorem*}{Theorem}
\AtBeginDocument{%
  \providecommand\BibTeX{{%
    \normalfont B\kern-0.5em{\scshape i\kern-0.25em b}\kern-0.8em\TeX}}}

\copyrightyear{2023}
\acmYear{2023}
\setcopyright{acmlicensed}\acmConference[KDD '23]{Proceedings of the 29th ACM SIGKDD Conference on Knowledge Discovery and Data Mining}{August 6--10, 2023}{Long Beach, CA, USA}
\acmBooktitle{Proceedings of the 29th ACM SIGKDD Conference on Knowledge Discovery and Data Mining (KDD '23), August 6--10, 2023, Long Beach, CA, USA}
\acmPrice{15.00}
\acmDOI{10.1145/3580305.3599295}
\acmISBN{979-8-4007-0103-0/23/08}

\begin{document}

\title{DCdetector: Dual Attention Contrastive Representation Learning \\ for Time Series Anomaly Detection}

\author{Yiyuan Yang}
\authornote{The first three authors contributed equally to this research.}
\authornote{Work done as an intern in DAMO Academy, Alibaba Group. He is now with the Department of Computer Science, University of Oxford, OX1 3SA, Oxford, UK.}
\affiliation{%
  \institution{University of Oxford}
  \city{Oxford}
  \country{UK}}
\email{yiyuan.yang@cs.ox.ac.uk}

\author{Chaoli Zhang}
\authornotemark[1]
\affiliation{%
  \institution{DAMO Academy, Alibaba Group}
  \city{Hangzhou}
  \country{China}
  }
 \email{chaoli.zcl@alibaba-inc.com}

\author{Tian Zhou}
\authornotemark[1]
\affiliation{%
  \institution{DAMO Academy, Alibaba Group}
  \city{Hangzhou}
  \country{China}
  }
 \email{tian.zt@alibaba-inc.com}

\author{Qingsong Wen}
\authornote{Corresponding author}
\affiliation{%
  \institution{DAMO Academy, Alibaba Group}
  \city{Bellevue}
  \country{USA}
  }
 \email{qingsong.wen@alibaba-inc.com}

\author{Liang Sun}
\affiliation{%
  \institution{DAMO Academy, Alibaba Group}
  \city{Bellevue}
  \country{USA}
  }
 \email{liang.sun@alibaba-inc.com}

\renewcommand{\shortauthors}{Yiyuan Yang, Chaoli Zhang, Tian Zhou, Qingsong Wen, \& Liang Sun}

\begin{abstract}

\end{abstract}



\begin{abstract}
    Time series anomaly detection is critical for a wide range of applications. It aims to identify deviant samples from the normal sample distribution in time series. The most fundamental challenge for this task is to learn a representation map that enables effective discrimination of anomalies. Reconstruction-based methods still dominate, but the representation learning with anomalies might hurt the performance with its large abnormal loss. On the other hand, contrastive learning aims to find a representation that can clearly distinguish any instance from the others, which can bring a more natural and promising representation for time series anomaly detection. In this paper, we propose DCdetector, a multi-scale dual attention contrastive representation learning model.
    DCdetector utilizes a novel dual attention asymmetric design to create the permutated environment and pure contrastive loss to guide the learning process, thus learning a permutation invariant representation with superior discrimination abilities. Extensive experiments show that DCdetector achieves state-of-the-art results on multiple time series anomaly detection benchmark datasets. Code is publicly available at this URL\footnote{\url{https://github.com/DAMO-DI-ML/KDD2023-DCdetector}}.
\end{abstract}

\maketitle

\section{Introduction}

Time series anomaly detection is widely used in real-world applications, including but not limited to industrial equipment status monitoring, financial fraud detection, fault diagnosis, and daily monitoring and maintenance of automobiles~\cite{cook2019anomaly,ren2019time,anandakrishnan2018anomaly,golmohammadi2015time,yang2021early}. 
With the rapid development of different sensors, large-scale time series data has been collected during the system's running time in many different applications~\cite{wen2022robust, li2021block, yang2023sgdp}. Effectively discovering abnormal patterns in systems is crucial to ensure security and avoid economic losses~\cite{yang2021pipeline}. For example, in the energy industry, detecting anomalies in wind turbine sensors in time helps to avoid catastrophic failure. In the financial industry, detecting fraud is essential for reducing pecuniary loss. 

However, it is challenging to discover abnormal patterns from a mass of complex time series. Firstly, it is still being determined what the anomalies will be like. Anomaly is also called outlier or novelty, which means observation unusual, irregular, inconsistent, unexpected, rare, faulty, or simply strange depending on the situation~\cite{ruff2021unifying}. Moreover, the typical situation is usually complex, which makes it harder to define what is unusual or unexpected. For instance, the wind turbine works in different patterns with different weather situations. Secondly, anomalies are usually rare, so it takes work to get labels~\cite{yang2021long}. Most supervised or semi-supervised methods fail to work given limited labeled training data. Third, anomaly detection models should consider temporal, multidimensional, and non-stationary features for time series data~\cite{yang2021pipeline2}. 
Multidimensionality describes that there is usually a dependence among dimensions in multivariate time series, and non-stationarity means the statistical features of time series are unstable. Specifically, temporal dependency means the adjacent points have latent dependence on each other. Although every point should be labeled as normal or abnormal, it is not reasonable to consider a single point as a sample.

Researchers have designed various time-series anomaly detection methods to deal with these challenges. They can be roughly classified as statistical, 
classic machine learning, and deep learning-based methods~\cite{blazquez2021review,ruff2021unifying}. Machine learning methods, especially deep learning-based methods, have succeeded greatly due to their powerful representation advantages. Most of the supervised and semi-supervised methods~\cite{zhang2022tfad,chauhan2015anomaly,chen2021learning,park2018multimodal,niu2020lstm,zhao2020multivariate} can not handle the challenge of limited labeled data, especially the anomalies are dynamic and new anomalies never observed before may occur. 
Unsupervised methods are popular without strict requirements on labeled data, including one class classification-based, probabilistic-based, distance-based, forecasting-based, reconstruction-based approaches~\cite{deng2021graph,zong2018deep,su2019robust,li2022learning,campos2021unsupervised,zamanzadeh2022deep,ruff2021unifying}. 

Reconstruction-based methods learn a model to reconstruct normal samples, and thereby the instances \emph{failing} be reconstructed by the learned model are anomalies. Such an approach is developing rapidly due to its power in handling complex data by combining it with different machine learning models and its interpretability that the instances behave unusually abnormally.
However, it is usually challenging to learn a well-reconstructed model for normal data without being obstructed by anomalies. The situation is even worse in time series anomaly detection as the number of anomalies is unknown, and normal and abnormal points may appear in one instance, making it harder to learn a clean, well-reconstructed model for normal points. 

Recently, contrastive representative learning has attracted attention due to its diverse design and outstanding performance in downstream tasks in the computer vision field~\cite{ye2019unsupervised,he2020momentum,chen2020simple,caron2020unsupervised}. However, the effectiveness of contrastive representative learning still needs to be explored in the time-series anomaly detection area. In this paper, we propose a \textbf{D}ual attention \textbf{C}ontrastive representation learning anomaly \textbf{detector} called \textbf{DCdetector} to handle the challenges in time series anomaly detection.
The key idea of our DCdetector is that normal time series points share the latent pattern, which means normal points have strong correlations with other points. In contrast, the anomalies do not (\emph{i.e.}, weak correlations with others). Learning consistent representations for anomalies from different views will be hard but easy for normal points. The primary motivation is that if normal and abnormal points' representations are distinguishable, we can detect anomalies without a highly qualified reconstruction model.

Specifically, we propose a contrastive structure with two branches and a dual attention module, and two branches share network weights. This model is trained based on the similarity of two branches, as normal points are the majority. The representation inconsistency of anomaly will be conspicuous. Thus, the representation difference between normal and abnormal data is enlarged without a highly qualified reconstruction model. 
To capture the temporal dependency in time series, DCdetector utilizes patching-based attention networks as the basic module. A multi-scale design is proposed to reduce information loss during patching. DCdetector takes all channels into representation efficiently with a channel independence design for multivariate time series. 
In particular, DCdetector does not require prior knowledge about anomalies and thus can handle new outliers never observed before. The main contributions of our DCdetector are summarized as follows:

\begin{itemize}
    \item Architecture: A contrastive learning-based dual-branch attention structure is designed to learn a permutation invariant representation that enlarges the representation differences between normal points and anomalies. Also, channel independence patching is proposed to enhance local semantic information in time series. Multi-scale is proposed in the attention module to reduce information loss during patching. 
    \item Optimization: An effective and robust loss function is designed based on the similarity of two branches. Note that the model is trained purely contrastively without reconstruction loss, which reduces distractions from anomalies. 
    \item Performance \& Justification: DCdetector achieves performance comparable or superior to state-of-the-art methods on seven multivariate and one univariate time series anomaly detection benchmark datasets. We also provide justification discussion to explain how our model avoids collapse without negative samples.

\end{itemize} 

\section{Related Work}
In this section, we show the related literature for this work. The relevant works include anomaly detection and contrastive representation learning.
\paragraph{Time Series Anomaly Detection}

\begin{figure*}[!ht]
    \includegraphics[width=1.0\textwidth]{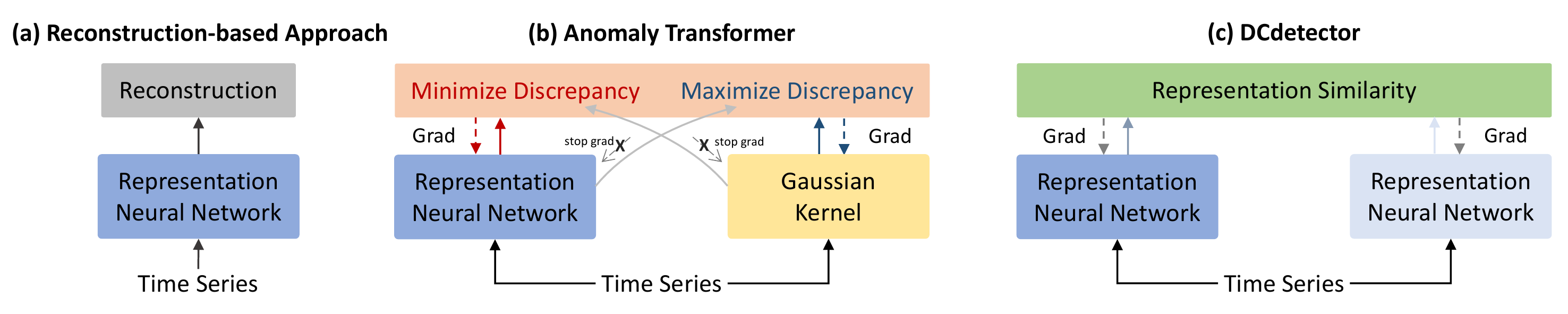}
    \caption{Architecture comparison of three approaches. The reconstruction-based approach uses a representation neural network to learn the pattern of normal points and do reconstruction. In Anomaly Transformer, the prior discrepancy is learned with Gaussian Kernel and the association discrepancy is learned with a transformer module; MinMax association learning is also critical and reconstruction loss is contained. DCdetector is concise without a specially designed Gaussian Kernel or a MinMax learning strategy, nor a reconstruction loss.}
    \label{fig:art-compare}
\end{figure*}

There are various approaches to detect anomalies in time series, including statistical methods, classical machine learning methods, and deep learning methods~\cite{schmidl2022anomaly}. Statistical methods include using moving averages, exponential smoothing~\cite{phillips2015business}, and the autoregressive integrated moving average (ARIMA) model~\cite{box1970distribution}. Machine learning methods include clustering algorithms such as k-means~\cite{kant2019time} and density-based methods, as well as classification algorithms such as decision trees~\cite{liu2008isolation,karczmarek2020k} and support vector machines (SVMs).
Deep learning methods include using autoencoders, variational autoencoders (VAEs)~\cite{sakurada2014anomaly,park2018multimodal}, and recurrent neural networks (RNNs)~\cite{canizo2019multi,su2019robust} such as long short-term memory (LSTM) networks~\cite{zamanzadeh2022deep}. Recent works in time series anomaly detection also include generative adversarial networks (GANs) based methods~\cite{li2019mad,zhou2019beatgan,chen2021daemon,chen2021daemon} and deep reinforcement learning (DRL) based methods~\cite{yu2020policy,huang2018towards}. In general, deep learning methods are more effective in identifying anomalies in time series data, especially when the data is high-dimensional or non-linear.

In another view, time series anomaly detection models can be roughly divided into two categories: supervised and unsupervised anomaly detection algorithms. Supervised methods can perform better when the anomaly label is available or affordable. Such methods can be dated back to AutoEncoder~\cite{sakurada2014anomaly}, LSTM-VAE~\cite{park2018multimodal}, Spectral Residual (SR)~\cite{ren2019time}, RobustTAD~\cite{jingkun20_TAD} and so on. On the other hand, an unsupervised anomaly detection algorithm can be applied in cases where the anomaly labels are difficult to obtain. Such versatility results in the community's long-lasting interest in developing new unsupervised time-series anomaly detection methods, including DAGMM~\cite{zong2018deep}, OmniAnomaly~\cite{su2019robust}, GDN~\cite{deng2021graph}, RDSSM~\cite{li2022learning} and so on. 
Unsupervised deep learning methods have been widely studied in time series anomaly detection. The main reasons are as follows. First, it is usually hard or unaffordable to get labels for all time series sequences in real-world applications. Second, deep models are powerful in representation learning and have the potential to get a decent detection accuracy under the unsupervised setting. Most of them are based on a reconstruction approach where a well-reconstructed model is learned for normal points; Then, the instances failing to be reconstructed are anomalies. 
Recently, some self-supervised learning-based methods have been proposed to enhance the generalization ability in unsupervised anomaly detection~\cite{zhao2020multivariate, jiao2022timeautoad, zhang2022adaptive}.

\paragraph{Contrastive Representation Learning}
The goal of contrastive representation learning is to learn an embedding space in which similar data samples stay close to each other while dissimilar ones are far apart. The idea of contrastive learning can be traced back to InstDic~\cite{wu2018unsupervised}. Classical contrastive models create <positive, negative> sample pairs to learn a representation where positive samples are near each other (pulled together) and far from negative samples (pushed apart)~\cite{ye2019unsupervised,he2020momentum,chen2020simple,caron2020unsupervised}. Their key designs are about how to define negative samples and deal with the high computation power/large batches requirements~\cite{khan2022contrastive}. On the other hand, BYOL~\cite{grill2020bootstrap} and SimSiam~\cite{chen2021exploring} get rid of negative samples involved, and such a simple siamese model (SimSiam) achieves comparable performance with other state-of-the-art complex architecture.

It is illuminating to make the distance of two-type samples larger using contrastive design. We try to distinguish time series anomalies and normal points with a well-designed multi-scale patching-based attention module. Moreover, our DCdetector is also free from negative samples and does not fall into a trivial solution even without the "stop gradient".

\section{Methodology}

Consider a multivariate time-series sequence of length T: $$ {\mathcal{X} = (x_1, x_2, \ldots, x_T),} $$

\noindent where each data point $x_t \in {\rm I\!R}^d$ is acquired at a certain timestamp $t$ from industrial sensors or machines, and $d$ is the data dimensionality, e.g., the number of sensors or machines. Our problem can be regarded as given input time-series sequence $\mathcal{X}$, for another unknown test sequence $\mathcal{X}_{test}$ of length $T'$ with the same modality as the training sequence, we want to predict $\mathcal{Y}_{test} = (y_1,y_2,\ldots,y_{T'})$. Here $y_t \in \{0,1\}$ where 1 denotes an anomalous data point and 0 denotes a normal data point.

As mentioned previously, representation learning is a powerful tool to handle the complex pattern of time series. Due to the high cost of gaining labels in practice, unsupervised and self-supervised methods are more popular. The critical issue in time series anomaly detection is to distinguish anomalies from normal points. Learning representations that demonstrate wide disparities without anomalies is promising. We amplify the advantages of contrastive representation learning with a dual attention structure. 

In some way, the underlining inductive bias we used here is similar to what Anomaly Transformer explored~\cite{xu2021anomaly}. That is, anomalies have less connection or interaction with the whole series than their adjacent points. The Anomaly Transformer detects anomalies by association discrepancy between a learned Gaussian kernel and attention weight distribution. In contrast, we proposed DCdetector, which achieves a similar goal in a much more general and concise way with a dual-attention self-supervised contrastive-type structure. 

To better position our work in the landscape of time series anomaly detection, we give a brief comparison of three approaches. To be noticed, Anomaly Transformer is a representation of a series of explicit association modeling works~\cite{cheng2009detection,Zhao2020MultivariateTA,Deng2021GraphNN,Boniol2020Series2GraphGS}, not implying it is the only one. We merely want to make a more direct comparison with the closest work here.  Figure~\ref{fig:art-compare} shows the architecture comparison of three approaches. The reconstruction-based approach (Figure~\ref{fig:art-compare}(a)) uses a representation neural network to learn the pattern of normal points and do reconstruction. Anomaly Transformer (Figure~\ref{fig:art-compare}(b)) takes advantage of the observation that it is difficult to build nontrivial associations from abnormal points to the whole series. Thereby, the prior discrepancy is learned with Gaussian Kernel and the association discrepancy is learned with a transformer module. MinMax association learning is also critical for Anomaly Transformer and reconstruction loss is contained. In contrast, the proposed DCdetector (Figure~\ref{fig:art-compare}(c)) is concise, in the sense that it does not need a specially designed Gaussian Kernel, a MinMax learning strategy, or a reconstruction loss. The DCdetector mainly leverages the designed contrastive learning-based dual-branch attention for discrepancy learning of anomalies in different views to enlarge the differences between anomalies and normal points. The simplicity and effectiveness contribute to DCdetector’s versatility.

\begin{figure*}[!t]
    \centering
    \includegraphics[width=1.0\textwidth]{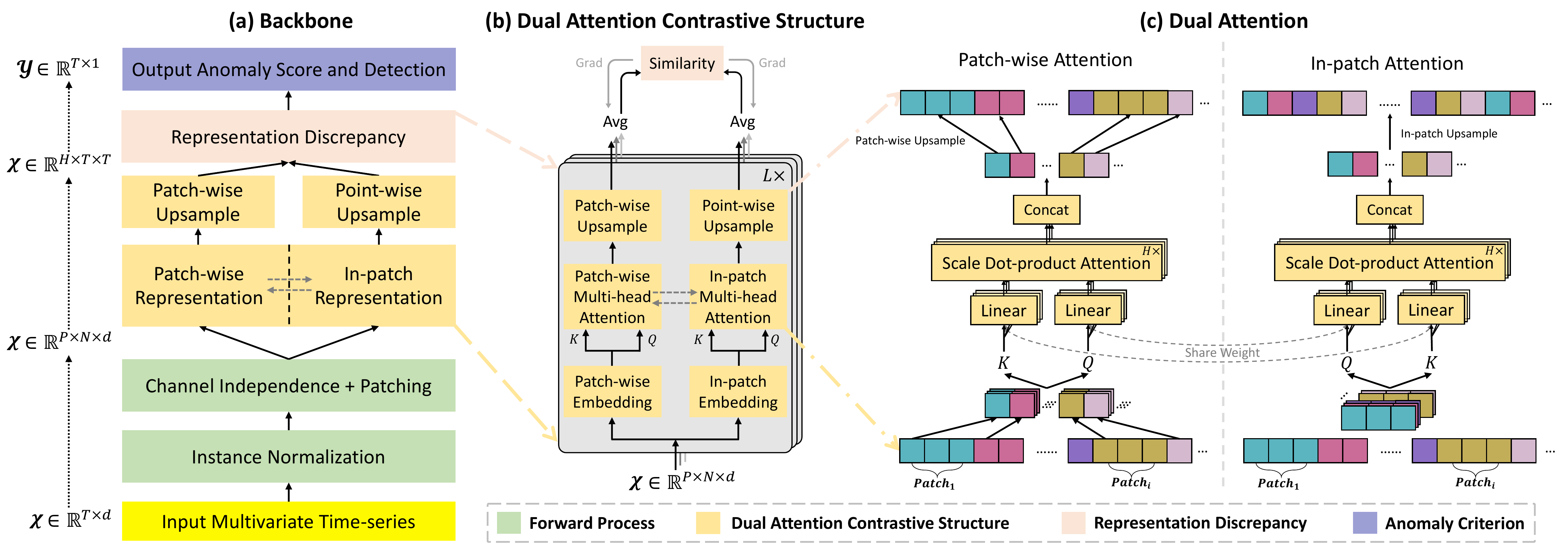}
    \caption{The workflow of the DCdetector framework. DCdetector consists of four main components: Forward Process module, Dual Attention Contrastive Structure module, Representation Discrepancy module, and Anomaly Criterion module.} 
    \label{fig:workflow}
\end{figure*}

\begin{figure*}[!t]
    \includegraphics[width=1.0\textwidth]{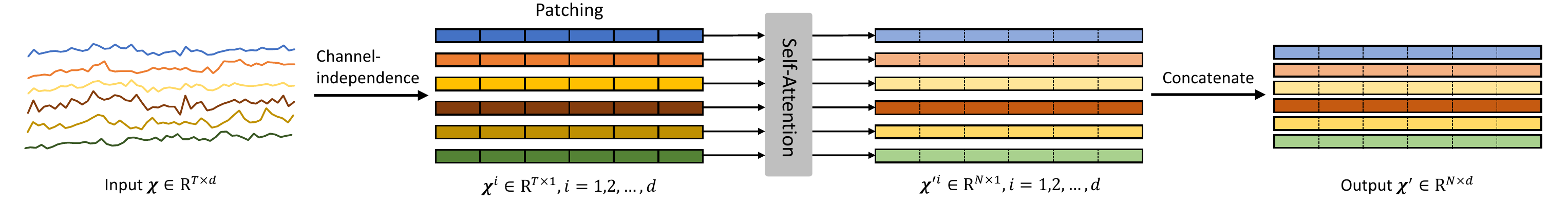}
    \caption{Basic patching attention with channel independence. Each channel in the multivariate time series input is considered as a single time series and divided into patches. Each channel shares the same self-attention network, and the representation results are concatenated as the final output.}
    \label{fig:patch}
\end{figure*}

\subsection{Overall Architecture}
Figure~\ref{fig:workflow} shows the overall architecture of the DCdetector, which consists of four main components, Forward Process module, Dual Attention Contrastive Structure module, Representation Discrepancy module, and Anomaly Criterion module.

The input multivariate time series in the Forward Process module is normalized by an instance normalization~\cite{kim2021reversible,ulyanov2017improved} module. The inputs to the instance normalization all come from the independent channels themselves. It can be seen as a consolidation and adjustment of global information, and a more stable approach to training processing. Channel independence assumption has been proven helpful in multivariate time series forecasting tasks~\cite{salinas2020deepar,li2019enhancing} to reduce parameter numbers and overfitting issues. Our DCdetector follows such channel independence setting to simplify the attention network with patching. 

More specifically, the basic patching attention with channel independence is shown in Figure~\ref{fig:patch}. Each channel in the multivariate time series input ($\mathcal{X}\in {\rm I\!R}^{T\times d}$) is considered as a single time series ($\mathcal{X}^{i}\in {\rm I\!R}^{T\times 1}, i = 1,2,\dots, d$) and divided into patches. Each channel shares the same self-attention network, and the representation results ($\mathcal{X}'^{i}\in {\rm I\!R}^{N\times 1}, i = 1,2,\dots, d$) is concatenated as the final output ($\mathcal{X}'\in {\rm I\!R}^{N\times d}$). 
In the implementation phase, running a sliding window in time series data is widely used in time series anomaly detection tasks~\cite{shen2020timeseries,xu2021anomaly} and has little influence on the main design. More implementation details are left in the experiment section.

The Dual Attention Contrastive Structure module is critical in our design. It learns the representation of inputs in different views. The insight is that, for normal points, most of them will share the same latent pattern even in different views (a strong correlation is not easy to be destroyed). However, as anomalies are rare and do not have explicit patterns, it is hard for them to share latent modes with normal points or among themselves (\emph{i.e.}, anomalies have a weak correlation with other points). Thus, the difference will be slight for normal point representations in different views and large for anomalies. We can distinguish anomalies from normal points with a well-designed Representation Discrepancy criterion. The details of Dual Attention Contrastive Structure and Representation Discrepancy are left in the following Section~\ref{sec:dacs} and Section~\ref{sec:rd}.

As for the Anomaly Criterion, we calculate anomaly scores based on the discrepancy between the two representations and use a prior threshold for anomaly detection. The details are left in Section~\ref{sec:ac}.

\subsection{Dual Attention Contrastive Structure}\label{sec:dacs}

In DCdetector, we propose a contrastive representation learning structure with dual attention to get the representations of input time series from different views. Concretely, with the patching operation, DCdetector takes patch-wise and in-patch representations as two views. Note that it differs from traditional contrastive learning, where original and augmented data are considered as two views of the original data. Moreover, DCdetector does not construct <positive, negative> pairs like the typical contrastive methods~\cite{wu2018unsupervised,he2020momentum}. Instead, its basic setting is similar to the contrastive methods only using positive samples~\cite{chen2021exploring,grill2020bootstrap}.

\subsubsection{Dual Attention}
As shown in Figure~\ref{fig:workflow}, input time series $\mathcal{X}\in{\rm I\!R}^{T\times d}$ are patched as $\mathcal{X}\in {\rm I\!R}^{P \times N \times d}$ where $P$ is the size of patches and $N$ is the number of patches. Then, we fuse the channel information with the batch dimension and the input size becomes $\mathcal{X}\in {\rm I\!R}^{P \times N}$. With such patched time series, DCdetector learns representation in patch-wise and in-patch views with self-attention networks. Our dual attention can be encoded in $L$ layers, so for simplicity, we only use one of these layers as an example.

For the patch-wise representation, a single patch is considered as a unit, and the dependencies among patches are modeled by a multi-head self-attention network (named patch-wise attention). In detail, an embedded operation will be applied in the patch\_size ($P$) dimension, and the shape of embedding is $\mathcal{X_N}\in {\rm I\!R}^{N \times d_{model}}$. Then, we adopt multi-head attention weights to calculate the patch-wise representation. Firstly, initialize the query and key: 
\begin{equation}\label{eq1}
    \mathcal{Q_N}_i, \mathcal{K_N}_i = \mathcal{W_{Q}}_i \mathcal{X_{N}}_i, \mathcal{W_{K}}_i \mathcal{X_{N}}_i \quad 1 \leq i \leq H,
\end{equation}
where $\mathcal{Q_{N}}_i, \mathcal{K_{N}}_i \in {\rm I\!R}^{N \times \frac{d_{model}}{H}}$ denote the query and key, respectively, $\mathcal{W_{Q}}_i, \mathcal{W_{K}}_i \in {\rm I\!R}^{\frac{d_{model}}{H} \times \frac{d_{model}}{H}}$ represent learnable parameter matrices of $\mathcal{Q_{N}}_i, \mathcal{K_{N}}_i$, and $H$ is the head number. Then, compute the attention weights: 
\begin{equation}\label{eq2}
    Attn_{\mathcal{N}_i} = Softmax(\frac{\mathcal{Q_{N}}_i \mathcal{K_{N}}_i^T}{\sqrt{\frac{d_{model}}{H}}}),
\end{equation}
where $Softmax(\cdot)$ function normalizes the attention weight. Finally, contact the multi-head and get the final patch-wise representation $Attn_\mathcal{N}$, which is: 
\begin{equation}\label{eq3}
    Attn_{\mathcal{N}} = \text{Concat}(Attn_{\mathcal{N}_1},\cdots, Attn_{\mathcal{N}_H}) W_\mathcal{N}^{O},
\end{equation}
where $W_\mathcal{N}^{O} \in {\rm I\!R}^{d_{model} \times d_{model}}$ is a learnable parameter matrix.

Similarly, for the in-patch representation, the dependencies of points in the same patch are gained by a multi-head self-attention network (called in-patch attention). Note that the patch-wise attention network shares weights with the in-patch attention network. Specifically, another embedded operation will be applied in the patch\_number ($N$) dimension, and the shape of embedding is $\mathcal{X_P}\in {\rm I\!R}^{P \times d_{model}}$. Then, we adopt multi-head attention weights to calculate the in-patch representation. First, initialize the query and key: 
\begin{equation}\label{eq4}
    \mathcal{Q_P}_i, \mathcal{K_P}_i = \mathcal{W_{Q}}_i \mathcal{X_{P}}_i, \mathcal{W_{K}}_i \mathcal{X_{P}}_i \quad 1 \leq i \leq H,
\end{equation}
where $\mathcal{Q_{P}}_i, \mathcal{K_{P}}_i \in {\rm I\!R}^{P \times \frac{d_{model}}{H}}$ denote the query and key, respectively, and $\mathcal{W_{Q}}_i, \mathcal{W_{K}}_i \in {\rm I\!R}^{\frac{d_{model}}{H} \times \frac{d_{model}}{H}}$ represent learnable parameter matrices of $\mathcal{Q_{P}}_i, \mathcal{K_{N}}_i$. Then, compute the attention weights: 
\begin{equation}\label{eq5}
    Attn_{\mathcal{P}_i} = Softmax(\frac{\mathcal{Q_{P}}_i \mathcal{K_{P}}_i^T}{\sqrt{\frac{d_{model}}{H}}}),
\end{equation}
where $Softmax(\cdot)$ function normalizes the attention weight. Finally, contact the multi-head and get the final in-patch representation $Attn_{\mathcal{P}}$, which is: 
\begin{equation}\label{eq6}
    Attn_\mathcal{P} = \text{Concat}(Attn_{\mathcal{P}_1},\cdots, Attn_{\mathcal{P}_H}) W_\mathcal{P}^{O},
\end{equation}
where $W_\mathcal{P}^{O} \in {\rm I\!R}^{d_{model} \times d_{model}}$ is a learnable parameter matrix. 

Note that the $\mathcal{W_{Q}}_i, \mathcal{W_{K}}_i$ are the shared weights within the in-patch attention representation network and patch-wise attention representation network.

\begin{figure}[t]
    \centering
    \includegraphics[width=1.0\columnwidth]{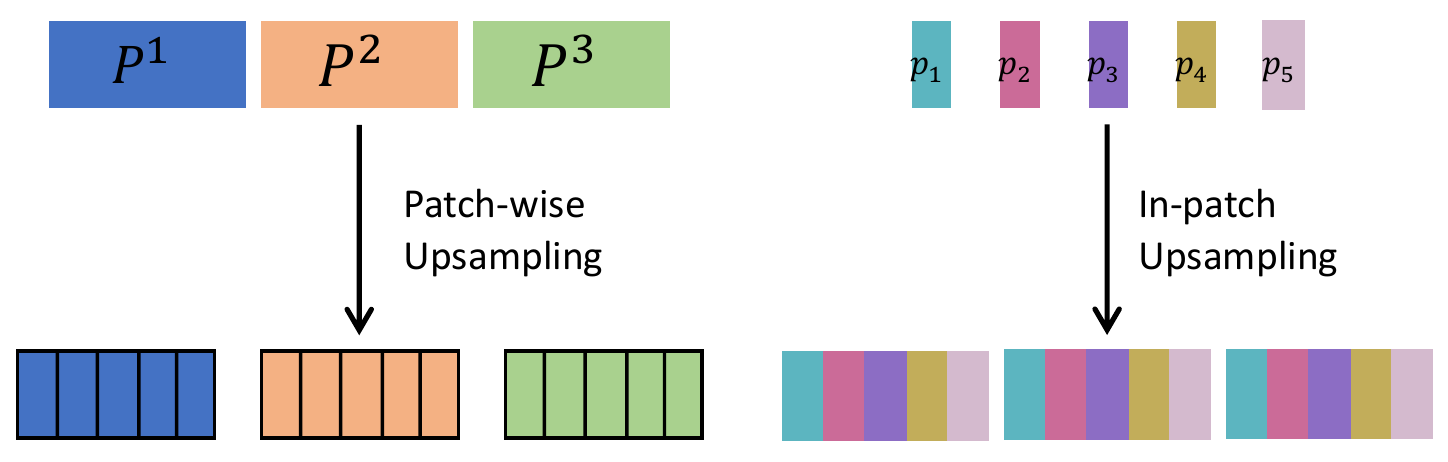}
    \caption{A simple example of how up-sampling is done. For patch-wise branch, repeating is done in patches (from patch to points). For in-patch branch, repeating is done from "one" patch to a full number of patches (from points to patches).}
    \label{fig:upsample}
\end{figure}
\subsubsection{Up-sampling and Multi-scale Design}
Although the patching design benefits from gaining local semantic information, patch-wise attention ignores the relevance among points in a patch, and in-patch attention ignores the relevance among patches. To compare the results of two representation networks, we need to do up-sampling first. For the patch-wise branch, as we only have the dependencies among patches, repeating is done inside patches (\emph{i.e.}, from patch to points) for up-sampling, and we will get the final patch-wise representation $\mathcal{N}$. For the in-patch branch, as only dependencies among patch points are gained, repeating is done from "one" patch to a full number of patches, and we will get the final in-patch representation $\mathcal{P}$. 

A simple example is shown in Figure~\ref{fig:upsample} where a patch is noted as $P^i$, and a point is noted as $p_i$. Such patching and repeating up-sampling operations inevitably lead to information loss. To keep the information from the original data better, DCdetector introduces a multi-scale design for patching representation and up-sampling. The final representation concatenates results in different scales (\emph{i.e.}, patch sizes). Specifically, we can preset a list of various patches to perform parallel patching and the computation of dual attention representations, simultaneously. After upsampling each patch part, they are summed to obtain the final patch-wise representation $\mathcal{N}$ and in-patch representation $\mathcal{P}$, which are
\begin{align}
     \mathcal{N} = \text{Upsampling}(Attn_{\mathcal{N}}),  \quad    \mathcal{P} = \text{Upsampling}(Attn_{\mathcal{P}}).
\end{align}

\subsubsection{Contrastive Structure}
Patch-wise and in-patch branches output representations of the same input time series in two different views. As shown in Figure \ref{fig:workflow} (c), patch-wise sample representation learns a weighted combination between sample points in the same position from each patch. In-patch sample representation, on the other hand, learns a weighted combination between points within the same patch. We can treat these two representations as permutated multi-view representations. The key inductive bias we exploit here is that normal points can maintain their representation under permutations while the anomalies can not. From such dual attention non-negative contrastive learning, we want to learn a \textbf{permutation invariant representation}. Learning details are left in Section~\ref{sec:rd}.

\subsection{Representation Discrepancy}\label{sec:rd}

With dual attention contrastive structure, representations from two views (patch-wise branch and in-patch branch) are gained. We formalize a loss function based on Kullback–Leibler divergence (KL divergence) to measure the similarity of such two representations. The intuition is that, as anomalies are rare and normal points share latent patterns, the same inputs' representations should be similar.  

\subsubsection{Loss function definition}
The loss function of the $\mathcal{P}$ and $\mathcal{N}$ can then be defined as 
\begin{equation}
    \mathcal{L_P}\{\mathcal{P}, \mathcal{N}; \mathcal{X}\} = \sum KL(\mathcal{P}, \text{Stopgrad}(\mathcal{N})) + KL(\text{Stopgrad}(\mathcal{N}), \mathcal{P}),
\end{equation}
\begin{equation}
    \mathcal{L_N}\{\mathcal{P}, \mathcal{N}; \mathcal{X}\} = \sum KL(\mathcal{N}, \text{Stopgrad}(\mathcal{P})) + KL(\text{Stopgrad}(\mathcal{P}), \mathcal{N}),
\end{equation}
where $\mathcal{X}$ is the input time series, $KL(\cdot||\cdot)$ is the KL divergence distance, $\mathcal{P}$ and $\mathcal{N}$ are the representation result matrices of the in-patch branch and the patch-wise branch, respectively. Stop-gradient (labeled as 'Stopgrad') operation is also used in our loss function to train two branches asynchronously. Then, the total loss function $\mathcal{L}$ is defined as
\begin{equation}
    \mathcal{L} = \frac{\mathcal{L_N}-\mathcal{L_P}}{len(\mathcal{N})}.
\end{equation}

Unlike most anomaly detection works based on reconstruction framework~\cite{ruff2021unifying}, DCdetector is a self-supervised framework based on representation learning, and no reconstruction part is utilized in our model. There is no doubt that reconstruction helps to detect the anomalies which behave not as expected. However, it is not easy to build a suitable encoder and decoder to 'reconstruct' the time series as they are expected to be with anomalies' interference. Moreover, the ability of representation is restricted as the latent pattern information is not fully considered. 

\subsubsection{Discussion about Model Collapse}
Interestingly, with only single-type inputs (or saying, no negative samples included), our DCdetector model does not fall into a trivial solution (model collapse). SimSiam~\cite{chen2021exploring} gives the main credit for avoiding model collapse to stop gradient operation in their setting. However, we find that DCdetector still works without stop gradient operation, although with the same parameters, the no stop gradient version does not gain the best performance. Details are shown in the ablation study (Section~\ref{sec:ablation}). 

A possible explanation is that our two branches are totally asymmetric. Following the unified perspective proposed in~\cite{zhang2022does}, consider the output vector of a branch as $Z$ and $Z$ can be decomposed into two parts $o$ and $r$ as $Z = o+r$, where $o = {\rm I\!E}[Z]$ is the center vector defined as an average of $Z$ in the whole representation space, 
and $r$ is the residual vector. When the collapse happens, all vectors $Z$ fall into the center vector $o$ and $o$ dominates over $r$. With two branches noted as $Z_p = o_p + r_p$, $Z_n = o_n + r_n$, if the branches are symmetric, \emph{i.e.}, $o_p = o_n$, then the distance between them is $Z_p - Z_n = r_p - r_n$. As $r_p$ and $r_n$ come from the same input example, it will lead to collapse. Fortunately, the two branches in DCdetector are asymmetric, so it is not easy for $o_p$ to be the same as $o_n$ even when $r_p$ and $r_n$ are similar. Thus, due to our asymmetric design, DCdetector is hard to fall into a trivial solution.

\subsection{Anomaly Criterion} \label{sec:ac}

With the insight that normal points usually share latent patterns (with strong correlation among them), thus the distances of representation results from different views for normal points are less than that for anomalies. The final anomaly score of $\mathcal{X}\in {\rm I\!R}^{T\times d}$ is defined as 
\begin{equation}
    \text{AnomalyScore}(\mathcal{X}) = \sum KL(\mathcal{P},\text{Stopgrad}(\mathcal{N}))+KL(\mathcal{N},\text{Stopgrad}(\mathcal{P})).
\end{equation}
It is a point-wise anomaly score, and anomalies result in higher scores than normal points. 

Based on the point-wise anomaly score, a hyperparameter threshold $\delta$ is used to decide if a point is an anomaly (1) or not (0). If the score exceeds the threshold, the output $\mathcal{Y}$ is an anomaly. That is 
\begin{equation} \label{eq11}
\mathcal{Y}_i = 
\begin{cases}
\text{1: anomaly} \quad \text{AnomalyScore}(\mathcal{X}_i) \geq \delta \\
\text{0: normal}  \quad \quad \!\!\! \text{AnomalyScore}(\mathcal{X}_i) < \delta .
\end{cases}
\end{equation}

\section{Experiments}

\subsection{Benchmark Datasets} \label{Chap: Benchmarks}
We adopt eight representative benchmarks from five real-world applications to evaluate DCdetector: (1) \textbf{MSL} (Mars Science Laboratory dataset) is collected by NASA and shows the condition of the sensors and actuator data from the Mars rover \cite{hundman2018detecting}. (2) \textbf{SMAP} (Soil Moisture Active Passive dataset) is also collected by NASA and presents the soil samples and telemetry information used by the Mars rover \cite{hundman2018detecting}. Compared with MSL, SMAP has more point anomalies.
(3) \textbf{PSM} (Pooled Server Metrics dataset) is a public dataset from eBay Server Machines with 25 dimensions \cite{abdulaal2021practical}. (4) \textbf{SMD} (Server Machine Dataset) is a five-week-long dataset collected from an internet company compute cluster, which stacks accessed traces of resource utilization of 28 machines \cite{su2019robust}. (5) \textbf{SWaT} (Secure Water Treatment) is a 51-dimension sensor-based dataset collected from critical infrastructure systems under continuous operations \cite{mathur2016swat}. (6) \textbf{NIPS-TS-SWAN} is an openly accessible comprehensive, multivariate time series benchmark extracted from solar photospheric vector magnetograms in Spaceweather HMI Active Region Patch series \cite{DVN/EBCFKM_2020,lai2021revisiting}. (7) \textbf{NIPS-TS-GECCO} is a drinking water quality dataset for the `internet of things', which is published in the 2018 genetic and evolutionary computation conference \cite{moritz2018gecco,lai2021revisiting}. Besides the above multivariate time series datasets, we also test univariate time series datasets. (8) \textbf{UCR} is provided by the Multi-dataset Time Series Anomaly Detection Competition of KDD2021, and contains 250 sub-datasets from various natural sources \cite{dau2019ucr,keogh2021multi}. It is a univariate time series of dataset subsequence anomalies. In each time series, there is one and only one anomaly. 
More details of the eight benchmark datasets are summarized in Table \ref{Tab: Dataset Description} in Appendix~\ref{sec:datasets}.

\begin{table*}[h!]
\caption{Overall results on real-world multivariate datasets. Performance ranked from lowest to highest. The \textit{P}, \textit{R} and \textit{F1} are the precision, recall and F1-score. All results are in \%, the best ones are in \textbf{Bold}, and the second ones are {\ul underlined}.}
\centering
\resizebox{1.0\textwidth}{!}{
\begin{tabular}{c|ccc|ccc|ccc|ccc|ccc}
\hline \hline
\textbf{Dataset} & \multicolumn{3}{c|}{\textbf{SMD}} & \multicolumn{3}{c|}{\textbf{MSL}} & \multicolumn{3}{c|}{\textbf{SMAP}} & \multicolumn{3}{c|}{\textbf{SWaT}} & \multicolumn{3}{c}{\textbf{PSM}} \\ \hline
\textbf{Metric} & \textbf{P} & \textbf{R} & \textbf{F1} & \textbf{P} & \textbf{R} & \textbf{F1} & \textbf{P} & \textbf{R} & \textbf{F1} & \textbf{P} & \textbf{R} & \textbf{F1} & \textbf{P} & \textbf{R} & \textbf{F1} \\ \hline
LOF & 56.34 & 39.86 & 46.68 & 47.72 & 85.25 & 61.18 & 58.93 & 56.33 & 57.60 & 72.15 & 65.43 & 68.62 & 57.89 & 90.49 & 70.61 \\
OCSVM & 44.34 & 76.72 & 56.19 & 59.78 & 86.87 & 70.82 & 53.85 & 59.07 & 56.34 & 45.39 & 49.22 & 47.23 & 62.75 & 80.89 & 70.67 \\
U-Time & 65.95 & 74.75 & 70.07 & 57.20 & 71.66 & 63.62 & 49.71 & 56.18 & 52.75 & 46.20 & 87.94 & 60.58 & 82.85 & 79.34 & 81.06 \\
IForest & 42.31 & 73.29 & 53.64 & 53.94 & 86.54 & 66.45 & 52.39 & 59.07 & 55.53 & 49.29 & 44.95 & 47.02 & 76.09 & 92.45 & 83.48 \\
DAGMM & 67.30 & 49.89 & 57.30 & 89.60 & 63.93 & 74.62 & 86.45 & 56.73 & 68.51 & {\ul 89.92} & 57.84 & 70.40 & 93.49 & 70.03 & 80.08 \\
ITAD & 86.22 & 73.71 & 79.48 & 69.44 & 84.09 & 76.07 & 82.42 & 66.89 & 73.85 & 63.13 & 52.08 & 57.08 & 72.80 & 64.02 & 68.13 \\
VAR & 78.35 & 70.26 & 74.08 & 74.68 & 81.42 & 77.90 & 81.38 & 53.88 & 64.83 & 81.59 & 60.29 & 69.34 & 90.71 & 83.82 & 87.13 \\
MMPCACD & 71.20 & 79.28 & 75.02 & 81.42 & 61.31 & 69.95 & 88.61 & 75.84 & 81.73 & 82.52 & 68.29 & 74.73 & 76.26 & 78.35 & 77.29 \\
CL-MPPCA & 82.36 & 76.07 & 79.09 & 73.71 & 88.54 & 80.44 & 86.13 & 63.16 & 72.88 & 76.78 & 81.50 & 79.07 & 56.02 & \textbf{99.93} & 71.80 \\
TS-CP2 & {\ul 87.42} & 66.25 & 75.38 & 86.45 & 68.48 & 76.42 & 87.65 & 83.18 & 85.36 & 81.23 & 74.10 & 77.50 & 82.67 & 78.16 & 80.35 \\
Deep-SVDD & 78.54 & 79.67 & 79.10 & {\ul 91.92} & 76.63 & 83.58 & 89.93 & 56.02 & 69.04 & 80.42 & 84.45 & 82.39 & 95.41 & 86.49 & 90.73 \\
BOCPD & 70.9 & 82.04 & 76.07 & 80.32 & 87.20 & 83.62 & 84.65 & 85.85 & 85.24 & 89.46 & 70.75 & 79.01 & 80.22 & 75.33 & 77.70 \\
LSTM-VAE & 75.76 & 90.08 & 82.30 & 85.49 & 79.94 & 82.62 & 92.20 & 67.75 & 78.10 & 76.00 & 89.50 & 82.20 & 73.62 & 89.92 & 80.96 \\
BeatGAN & 72.90 & 84.09 & 78.10 & 89.75 & 85.42 & 87.53 & 92.38 & 55.85 & 69.61 & 64.01 & 87.46 & 73.92 & 90.30 & 93.84 & 92.04 \\
LSTM & 78.55 & 85.28 & 81.78 & 85.45 & 82.50 & 83.95 & 89.41 & 78.13 & 83.39 & 86.15 & 83.27 & 84.69 & 76.93 & 89.64 & 82.80 \\
OmniAnomaly & 83.68 & 86.82 & 85.22 & 89.02 & 86.37 & 87.67 & 92.49 & 81.99 & 86.92 & 81.42 & 84.30 & 82.83 & 88.39 & 74.46 & 80.83 \\
InterFusion & 87.02 & 85.43 & 86.22 & 81.28 & 92.70 & 86.62 & 89.77 & 88.52 & 89.14 & 80.59 & 85.58 & 83.01 & 83.61 & 83.45 & 83.52 \\
THOC & 79.76 & 90.95 & 84.99 & 88.45 & 90.97 & 89.69 & 92.06 & 89.34 & 90.68 & 83.94 & 86.36 & 85.13 & 88.14 & 90.99 & 89.54 \\
AnomalyTrans & \textbf{88.47} & \textbf{92.28} & \textbf{90.33} & {\ul 91.92} & {\ul 96.03} & {\ul 93.93} & {\ul 93.59} & \textbf{99.41} & {\ul 96.41} & 89.10 & {\ul99.28} & {\ul94.22} & {\ul 96.94} & 97.81 & {\ul 97.37} \\ \hline
DCdetector & 83.59 & {\ul 91.10} & {\ul 87.18} & \textbf{93.69} & \textbf{99.69} & \textbf{96.60} & \textbf{95.63} & {\ul 98.92} & \textbf{97.02} & \textbf{93.11} & \textbf{99.77} & \textbf{96.33} & \textbf{97.14} & {\ul 98.74} & \textbf{97.94} \\ \hline \hline
\end{tabular}}
\label{Tab: Overall result}
\end{table*}
\begin{table*}[]
\caption{\small{Multi-metrics results on real-world multivariate datasets. Aff-P and Aff-R are the precision and recall of affiliation metric~\cite{huet2022local}, respectively. R\_A\_R and R\_A\_P are Range-AUC-ROC and Range-AUC-PR~\cite{paparrizos2022volume}, which denote two scores based on label transformation under ROC curve and PR curve, respectively. V\_ROC and V\_RR are volumes under the surfaces created based on ROC curve and PR curve~\cite{paparrizos2022volume}, respectively. All results are in \%, and the best ones are in \textbf{Bold}. } }
\label{Tab: multi-matrix results} 
\resizebox{1.0\textwidth}{!}
{
\begin{tabular}{c|c|cccccccc}
\hline \hline
\textbf{Dataset}               & \textbf{Method}     & \textbf{Acc} & \textbf{F1} & \textbf{Aff-P~\cite{huet2022local}} & \textbf{Aff-R~\cite{huet2022local}} & \textbf{R\_A\_R~\cite{paparrizos2022volume}} & \textbf{R\_A\_P~\cite{paparrizos2022volume}} & \textbf{V\_ROC~\cite{paparrizos2022volume}} & \textbf{V\_PR~\cite{paparrizos2022volume}} \\ \hline
\multirow{2}{*}{\textbf{MSL}}  & AnomalyTrans & 98.69             & 93.93             & 51.76                  & 95.98               & 90.04                & 87.87               & 88.20             & 86.26            \\
                               & DCdetector                & \textbf{99.06}    & \textbf{96.60}    & \textbf{51.84}         & \textbf{97.39}      & \textbf{93.17}       & \textbf{91.64}      & \textbf{93.15}    & \textbf{91.66}   \\ \hline
\multirow{2}{*}{\textbf{SMAP}} & AnomalyTrans & 99.05             & 96.41             & 51.39                  & \textbf{98.68}      & \textbf{96.32}       & 94.07               & \textbf{95.52}    & 93.37            \\
                               & DCdetector                & \textbf{99.21}    & \textbf{97.02}    & \textbf{51.46}         & 98.64               & 96.03                & \textbf{94.18}      & 95.19             & \textbf{93.46}   \\ \hline
                               
\multirow{2}{*}{\textbf{SWaT}}  & AnomalyTrans &   98.51     &  94.22      &   \textbf{53.03}   &    \textbf{98.08}          &    \textbf{97.89}  &  93.47     &\textbf{97.92}     & 93.49   \\
                               & DCdetector                & \textbf{99.09}  & \textbf{96.33}    &  52.40                &  97.67      &   96.63      & \textbf{94.06}        &  96.95        & \textbf{94.34}          \\ \hline
    
\multirow{2}{*}{\textbf{PSM}}  & AnomalyTrans & 98.68             & 97.37             & \textbf{55.35}         & 80.28               & \textbf{91.83}       & \textbf{93.03}      & \textbf{88.71}    & \textbf{90.71}   \\
                               & DCdetector                & \textbf{98.95}    & \textbf{97.94}    & 54.71                  & \textbf{82.93}      & 91.55                & 92.93               & 88.41             & 90.58            \\ \hline \hline
\end{tabular}
}
\end{table*}
\begin{table}[htbp]
\caption{Overall results on NIPS-TS datasets. Performance ranked from lowest to highest. All results are in \%, the best ones are in bold, and the second ones are {\ul underlined}.}
\resizebox{1.0\columnwidth}{!}
{
\begin{tabular}{c|ccc|ccc}
\hline \hline
\textbf{Dataset}    & \multicolumn{3}{c|}{\textbf{NIPS-TS-GECCO}} & \multicolumn{3}{c}{\textbf{NIPS-TS-SWAN}} \\ \hline
\textbf{Metric}     & \textbf{P}   & \textbf{R}   & \textbf{F1}   & \textbf{P}   & \textbf{R}  & \textbf{F1}  \\ \hline
OCSVM*              & 2.1          & 34.1         & 4.0           & 19.3         & 0.1         & 0.1          \\
MatrixProfile       & 4.6          & 18.5         & 7.4           & 16.7         & 17.5        & 17.1         \\
GBRT                & 17.5         & 14.0         & 15.6          & 44.7         & 37.5        & 40.8         \\
LSTM-RNN            & 34.3         & 27.5         & 30.5          & 52.7         & 22.1        & 31.2         \\
Autoregression      & 39.2         & 31.4         & 34.9          & 42.1         & 35.4        & 38.5         \\
OCSVM               & 18.5         & \textbf{74.3}         & 29.6          & 47.4         & 49.8        & 48.5         \\
IForest*            & 39.2         & 31.5         & 39.0          & 40.6         & 42.5        & 41.6         \\
AutoEncoder         & {\ul 42.4}         & 34.0         & 37.7          & 49.7         & 52.2        & 50.9         \\
AnomalyTrans        & 25.7         & 28.5         & 27.0          & {\ul 90.7}         & 47.4        & {\ul 62.3}        \\
IForest             & \textbf{43.9}         & 35.3         & {\ul 39.1}          & 56.9         & \textbf{59.8}        & 58.3         \\ \hline
DCdetector                & 38.3         & {\ul 59.7}         & \textbf{46.6}          & \textbf{95.5}         & {\ul 59.6}        & \textbf{73.4}        \\ \hline \hline
\end{tabular}
}
\label{Tab: NIPS-TS overall results}
\end{table}
\begin{table}[h!]
\caption{Overall results on univariate dataset. Results are in \%, and the best ones are in Bold.}
\centering
\resizebox{0.9\columnwidth}{!}{
\begin{tabular}{c|ccccc}
\hline \hline
\textbf{Dataset}    & \multicolumn{5}{c}{\textbf{UCR}}  \\  \hline
\textbf{Metric}     & \multicolumn{1}{c}{\textbf{Acc}} & \multicolumn{1}{c}{\textbf{P}} & \multicolumn{1}{c}{\textbf{R}} & \multicolumn{1}{c}{\textbf{F1}} & \multicolumn{1}{c}{\textbf{Count}} \\  \hline
AnomalyTrans & 99.49                            & 60.41                          & \textbf{100}                            & 73.08                           & 42    \\
DCdetector                & \textbf{99.51}                            & \textbf{61.62}                          & \textbf{100 }                           & \textbf{74.05}                           & \textbf{46}                 \\ \hline \hline
\end{tabular}}
\label{Tab: UCR and UCR_AUG results}
\end{table}
\begin{table*}[h!]
\caption{\small{Multi-metrics results on NIPS-TS datasets. All results are in \%, and the best ones are in \textbf{Bold}.}}
\resizebox{1.0\textwidth}{!}{
\begin{tabular}{c|c|cccccccccc}
\hline \hline
\textbf{Dataset}                      & \textbf{Method}     & \textbf{Acc}   & \textbf{P}     & \textbf{R}     & \textbf{F1}    & \textbf{Aff-P} & \textbf{Aff-R} & \textbf{R\_A\_R} & \textbf{R\_A\_P} & \textbf{V\_ROC} & \textbf{V\_PR} \\ \hline
\multirow{2}{*}{\textbf{NIPS-TS-SWAN}}  & AnomalyTrans & 84.57          & 90.71          & 47.43          & 62.29          & \textbf{58.45} & \textbf{9.49}  & 86.42                & 93.26               & 84.81             & 92.00            \\
                                     & DCdetector                & \textbf{85.94} & \textbf{95.48} & \textbf{59.55} & \textbf{73.35} & 50.48          & 5.63           & \textbf{88.06}       & \textbf{94.71}      & \textbf{86.25}    & \textbf{93.50}   \\ \hline
\multirow{2}{*}{\textbf{NIPS-TS-GECCO}} & AnomalyTrans & 98.03          & 25.65          & 28.48          & 26.99          & 49.23          & 81.20           & 56.35                & 22.53               & 55.45             & 21.71            \\
                                     & DCdetector                & \textbf{98.56} & \textbf{38.25} & \textbf{59.73} & \textbf{46.63} & \textbf{50.05} & \textbf{88.55} & \textbf{62.95}       & \textbf{34.17}      & \textbf{62.41}    & \textbf{33.67}   \\ \hline \hline
\end{tabular}
}
\label{Tab: NIPS-TS multi-matrix results}
\end{table*}

\subsection{Baselines and Evaluation Criteria} \label{Chap: Baselines}
We compare our model with the 26 baselines for comprehensive evaluations, including the reconstruction-based model: AutoEncoder \cite{sakurada2014anomaly}, LSTM-VAE \cite{park2018multimodal}, OmniAnomaly \cite{su2019robust}, BeatGAN \cite{zhou2019beatgan}, InterFusion \cite{li2021multivariate}, Anomaly Transformer \cite{xu2021anomaly}; the autoregression-based models: VAR \cite{anderson1976time}, Autoregression \cite{rousseeuw2005robust}, LSTM-RNN \cite{bontemps2016collective}, LSTM \cite{hundman2018detecting}, CL-MPPCA \cite{tariq2019detecting}; the density-estimation models: LOF \cite{breunig2000lof}, MPPCACD \cite{yairi2017data}, DAGMM \cite{zong2018deep}; the clustering-based methods: Deep-SVDD \cite{ruff2018deep}, THOC \cite{shen2020timeseries}, ITAD \cite{shin2020itad}; the classic methods: OCSVM \cite{tax2004support}, OCSVM-based subsequence clustering (OCSVM*), IForest \cite{liu2008isolation}, IForest-based subsequence clustering (IForest*), Gradient boosting regression (GBRT) \cite{elsayed2021we}; the change point detection and time series segmentation methods: BOCPD \cite{adams2007bayesian}, U-Time \cite{perslev2019u}, TS-CP2 \cite{deldari2021time}. We also compare our model with a time-series subsequence anomaly detection algorithm Matrix Profile \cite{yeh2016matrix}.

Besides, we adopt various evaluation criteria for comprehensive comparison, including the commonly-used evaluation measures: accuracy, precision, recall, F1-score; the recently proposed evaluation measures: affiliation precision/recall pair \cite{huet2022local} and Volume under the surface (VUS) \cite{paparrizos2022volume}. F1-score is the most widely used metric but does not consider anomaly events. Affiliation precision and recall are calculated based on
the distance between ground truth and prediction events. VUS metric takes anomaly events into consideration based on the receiver operator characteristic (ROC) curve. Different metrics provide different evaluation views. We employ the commonly-used adjustment technique for a fair comparison \cite{xu2018unsupervised, su2019robust, shen2020timeseries, xu2021anomaly}, according to which all abnormalities in an abnormal segment are considered to have been detected if a single time point in an abnormal segment is identified.

\subsection{Implementation Details} \label{Chap: Implementation}

We summarize all the default hyper-parameters as follows in our implementation. Our DCdetector model contains three encoder layers ($L=3$). The dimension of the hidden state $d_{model}$ is 256, and the number of attention heads $H$ is 1 for simplicity. We select various patch size and window size options for different datasets, as shown in Table \ref{Tab: Dataset Description} in Appendix~\ref{sec:datasets}. Our model defines an anomaly as a time point whose anomaly score exceeds a hyperparameter threshold $\delta$, and its default value to 1. For all experiments on the above hyperparameter selection and trade-off, please refer to Appendix~\ref{sec:appendix_extra_ablation}. 
Besides, all the experiments are implemented in PyTorch \cite{paszke2019pytorch} with one NVIDIA Tesla-V100 32GB GPU. Adam \cite{kingma2014adam} with default parameter is applied for optimization. We set the initial learning rate to $10^{-4}$ and the batch size to 128 with 3 epochs for all datasets.

\begin{table*}[!t]
\caption{Ablation studies on Stop Gradient in DCdetector. All results are in \%, and the best ones are in \textbf{Bold}.}
\resizebox{0.86\textwidth}{!}{
\begin{tabular}{cc|lll|lll|lll}
\hline \hline
\multicolumn{2}{c|}{\textbf{Stop Gradient}}          & \multicolumn{3}{c|}{\textbf{MSL}}                                                                  & \multicolumn{3}{c|}{\textbf{SMAP}}                                                                 & \multicolumn{3}{c}{\textbf{PSM}}                                                                  \\ \hline
\textbf{Patch-wise Branch} & \textbf{In-patch Branch} & \multicolumn{1}{c}{\textbf{P}} & \multicolumn{1}{c}{\textbf{R}} & \multicolumn{1}{c|}{\textbf{F1}} & \multicolumn{1}{c}{\textbf{P}} & \multicolumn{1}{c}{\textbf{R}} & \multicolumn{1}{c|}{\textbf{F1}} & \multicolumn{1}{c}{\textbf{P}} & \multicolumn{1}{c}{\textbf{R}} & \multicolumn{1}{c}{\textbf{F1}} \\ \hline
\ding{56}                         & \ding{56}                        & 91.99                          & 89.98                          & 90.97                           & 94.49                          & 96.56                          & 95.51                           & 96.86                          & 97.51                          & 97.18                           \\
\ding{52}                          & \ding{56}                       & 91.27                          & 72.61                          & 80.88                           & 94.46                             & 93.17                             & 93.81                               & 97.15                             & 98.51                           & 97.83                               \\
\ding{56}                         & \ding{52}                         & 92.18                          & 96.27                          & 94.18                           & 94.37                          & 98.19                          & 96.24                           & 96.98                          & 98.04                          & 97.51                           \\
\ding{52}                           & \ding{52}                         & 93.69      & 99.69      & \textbf{96.60}       & 95.63     & 98.92      & \textbf{97.02}       & 97.14      & 98.74      & \textbf{97.94}      \\ \hline \hline
\end{tabular}}
\label{Tab: Ablation Stop Gradient}
\end{table*}
\begin{table*}[!t]
\caption{Ablation studies on Forward Process module in DCdetector. All results are in \%, and the best ones are in \textbf{Bold}.}
\resizebox{0.85\textwidth}{!}{
\begin{tabular}{cc|ccc|ccc|ccc}
\hline \hline
\multicolumn{2}{c|}{\textbf{Forward Process}}     & \multicolumn{3}{c|}{\textbf{MSL}}      & \multicolumn{3}{c|}{\textbf{SMAP}}     & \multicolumn{3}{c}{\textbf{PSM}}      \\ \hline
\textbf{Bilateral Filter} & \textbf{Instance Norm} & \textbf{P} & \textbf{R} & \textbf{F1} & \textbf{P} & \textbf{R} & \textbf{F1} & \textbf{P} & \textbf{R} & \textbf{F1} \\ \hline
\ding{56}                         & \ding{56}                      & 92.58      & 96.68      & 94.59       & 94.65      & 97.38      & 96.00       & 97.01      & 97.79      & 97.40       \\
\ding{52}                         & \ding{56}                      & 92.64      & 98.74      & 95.59       & 94.48      & 98.48      & 96.44       & 97.11      & 98.44      & 97.77       \\
\ding{56}                         & \ding{52}                      & 93.69      & 99.69      & \textbf{96.60}       & 95.63      & 98.92      & \textbf{97.02}       & 97.14      & 98.74      & \textbf{97.94}       \\
\ding{52}                         & \ding{52}                      & 92.28      & 98.82      & 95.44       & 95.11      & 97.06      & 96.08       & 96.88      & 97.82      & 97.35      \\ \hline \hline
\end{tabular}}
\label{Tab: Ablation Preprocess results}
\end{table*}

\subsection{Main Results} \label{Chap: Main Results}

\subsubsection{Multivariate Anomaly Detection}

We first evaluate our DCdetector with nineteen competitive baselines on five real-world multivariate datasets as shown in Table \ref{Tab: Overall result}. 
It can be seen that our proposed DCdetector achieves SOTA results under the widely used F1 metric~\cite{ren2019time,xu2021anomaly} in most benchmark datasets.
It is worth mentioning that it has been an intense discussion among recent studies about how to evaluate the performance of anomaly detection algorithms fairly. Precision, Recall, and F1 score are still the most widely used metrics for comparison. Some additional metrics (affiliation precision/recall pair, VUS, etc.) are proposed to complement their deficiencies \cite{huet2022local,paparrizos2022volume,xu2018unsupervised, su2019robust, shen2020timeseries, xu2021anomaly}. To judge which metric is the best beyond the scope of our work, we include all the metrics here. 
As the recent Anomaly Transformer achieves better results than other baseline models, we mainly evaluate DCdetector with the Anomaly Transformer in this multi-metrics comparison as shown in Table~\ref{Tab: multi-matrix results}. It can be seen that DCdetector performs better or at least comparable with the Anomaly Transformer in most metrics.

We also evaluate the performance of another two datasets NIPS-TS-SWAN and NIPS-TS-GECCO in Table \ref{Tab: NIPS-TS overall results}, which are more challenging with more types of anomalies than the above five datasets. Although the two datasets have the highest (32.6\% in NIPS-TS-SWAN) and lowest (1.1\% in NIPS-TS-GECCO) anomaly ratio, DCdetector is still able to achieve SOTA results and completely outperform other methods.  
Similarly, multi-metrics comparisons between DCdetector and Anomaly Transformer are conducted and summarized in Table~\ref{Tab: NIPS-TS multi-matrix results}, and DCdetector still achieves better performance in most metrics.

\subsubsection{Univariate Anomaly Detection}
In this part, we compare the performance of the DCdetector and Anomaly Transformer in univariate time series anomaly detection.
We trained and tested separately for each of the sub-datasets in UCR datasets, and the average results are shown in Table \ref{Tab: UCR and UCR_AUG results}. The count indicates how many sub-datasets have reached SOTA. The sub-datasets of the UCR all have only one segment of subsequence anomalies, and DCdetecter can identify and locate them correctly and achieve optimal results.

\subsection{Model Analysis}

\begin{figure*}[htbp]
\centering
\includegraphics[width=1.0\textwidth]{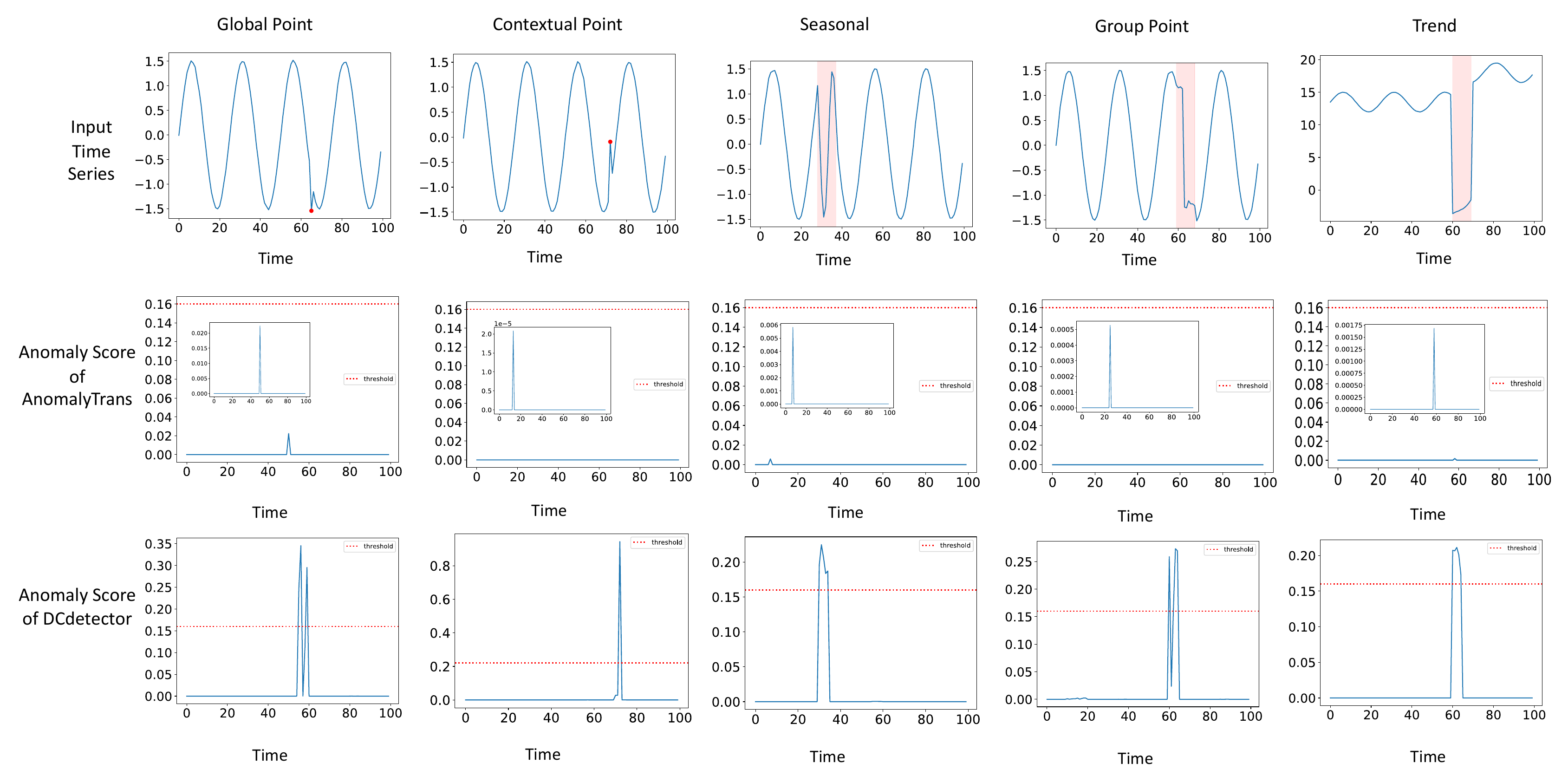}
\caption{Visualization comparisons of ground-truth anomalies and anomaly scores between DCdetector and Anomaly Transformer for different types of anomalies.} 
\label{fig:case2} 
\end{figure*}
\begin{figure*}[htbp]
	\centering
\subfigure[Window size]{\includegraphics[width=.195\textwidth]{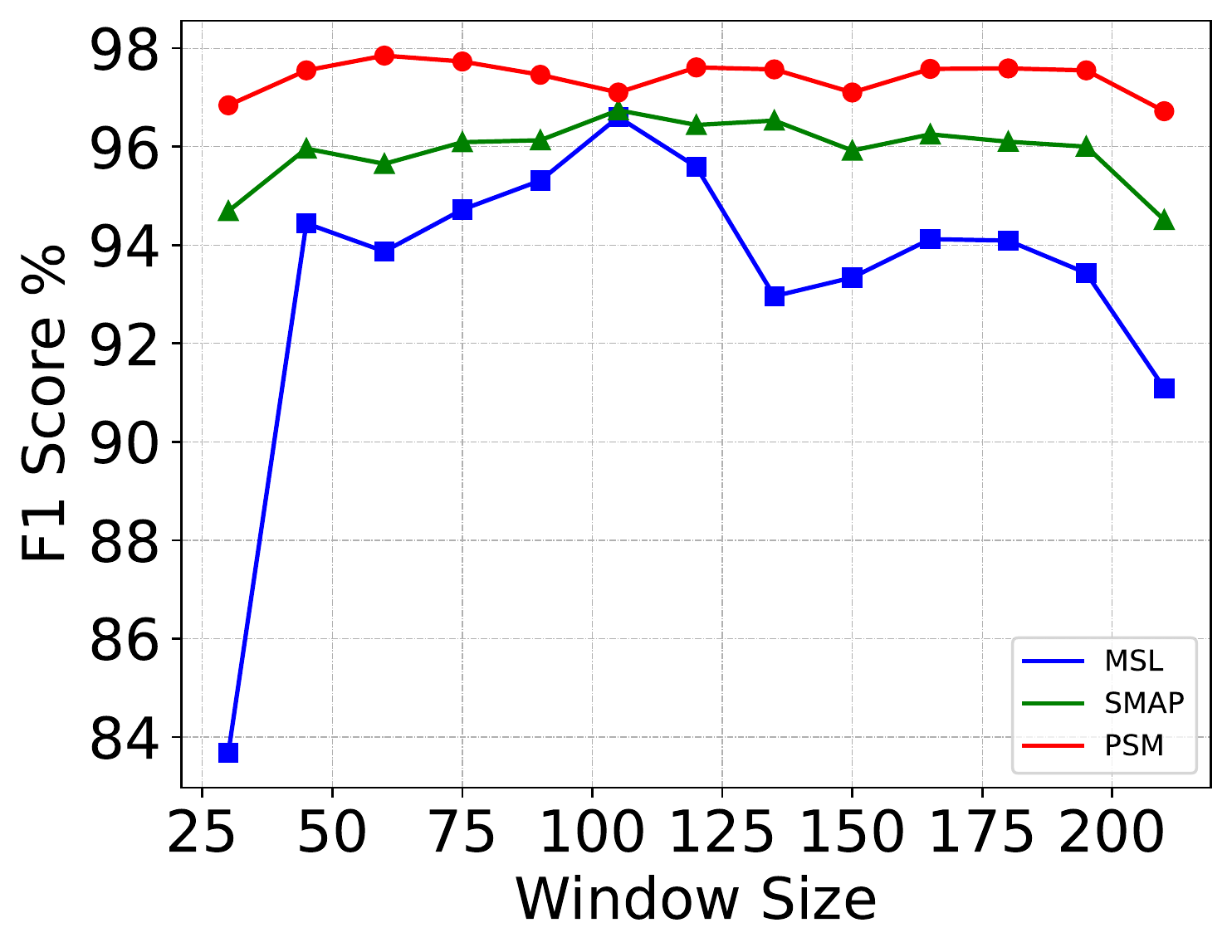}\label{fig:ab-win}}
	\subfigure[Multi-scale size]{\includegraphics[width=.195\textwidth]{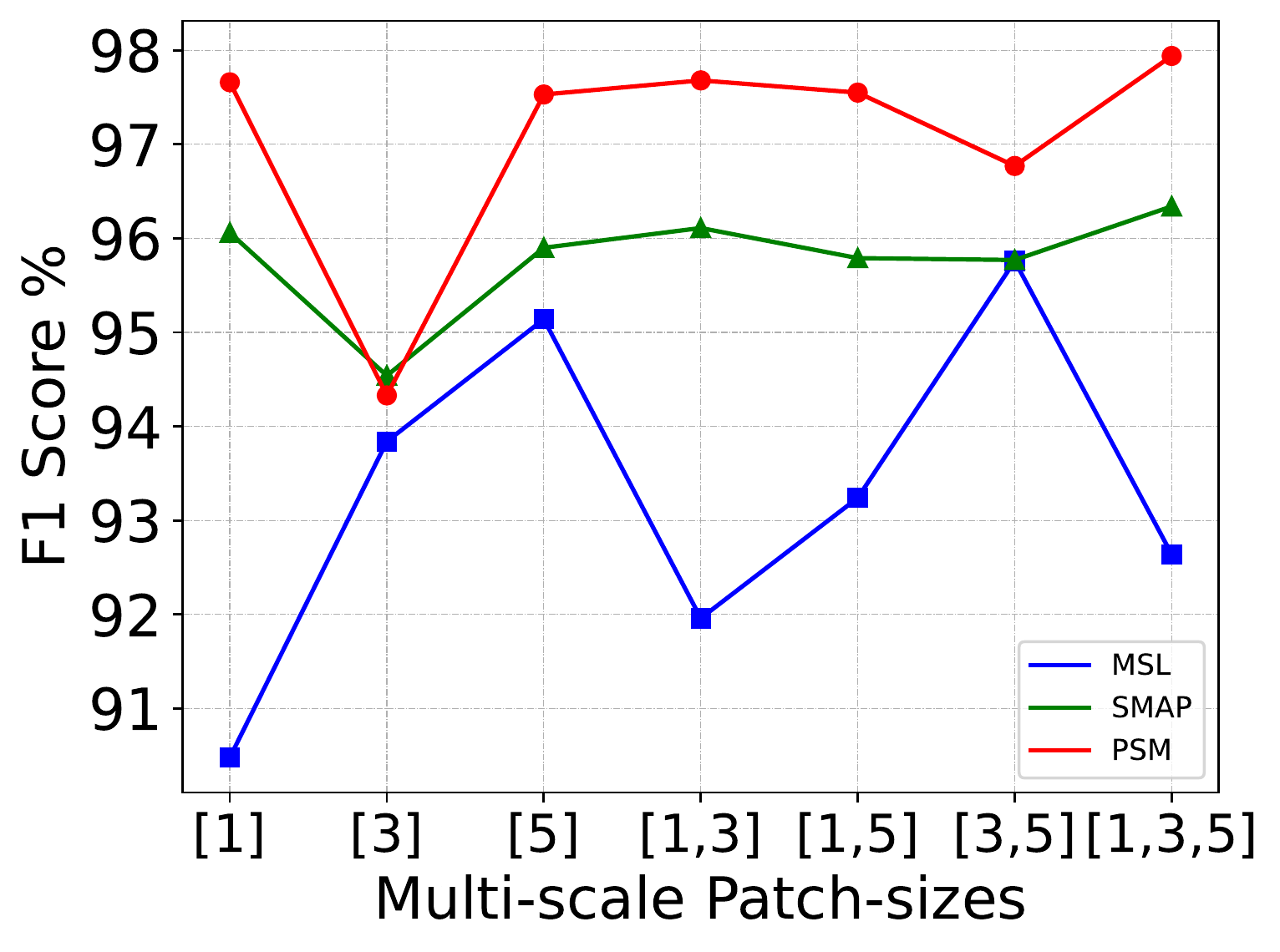}\label{fig:ab-scale}}
        \subfigure[Encoder layer number]{\includegraphics[width=.195\textwidth]{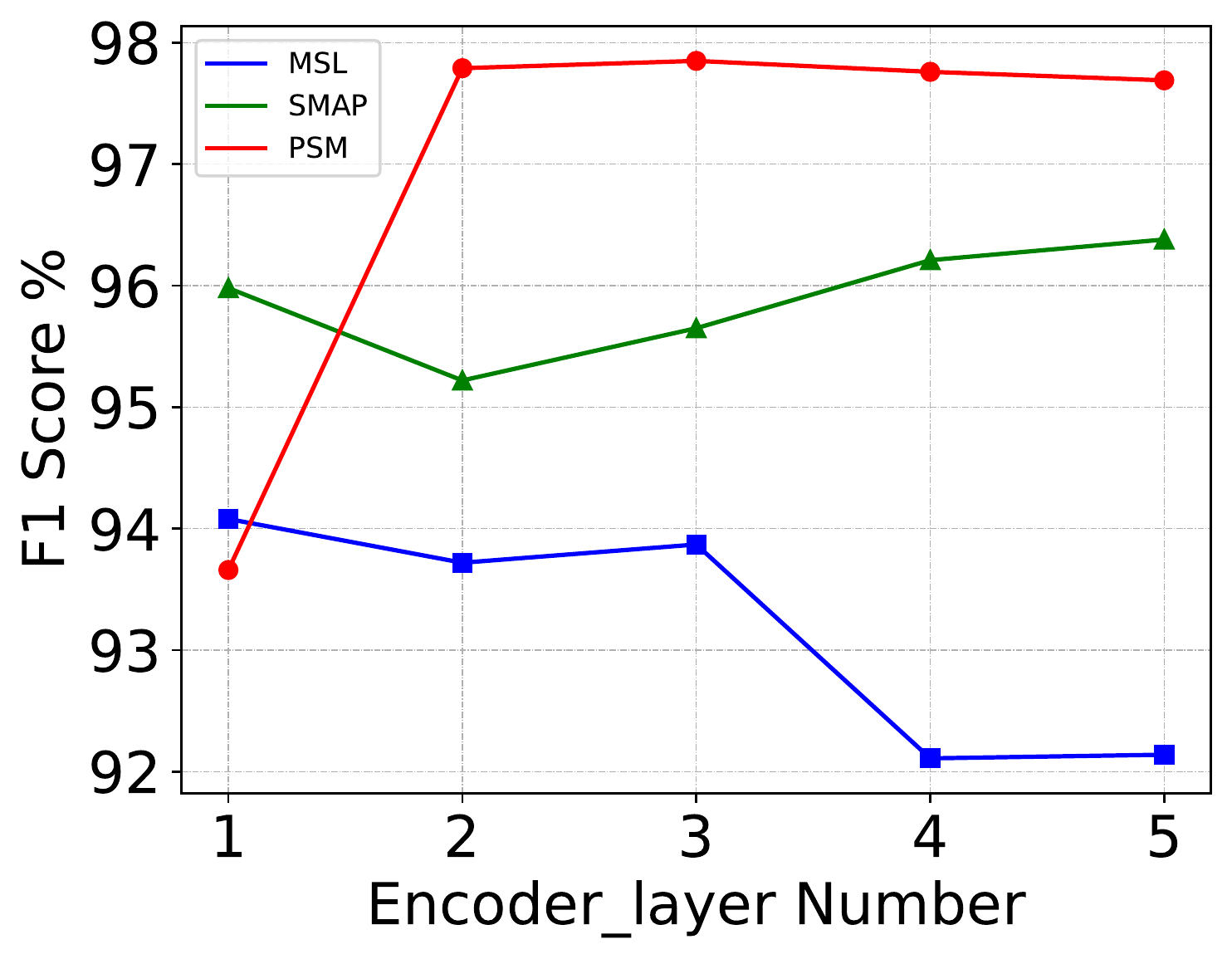}\label{fig:ab-layer}}
	\subfigure[Attention head number]{\includegraphics[width=.195\textwidth]{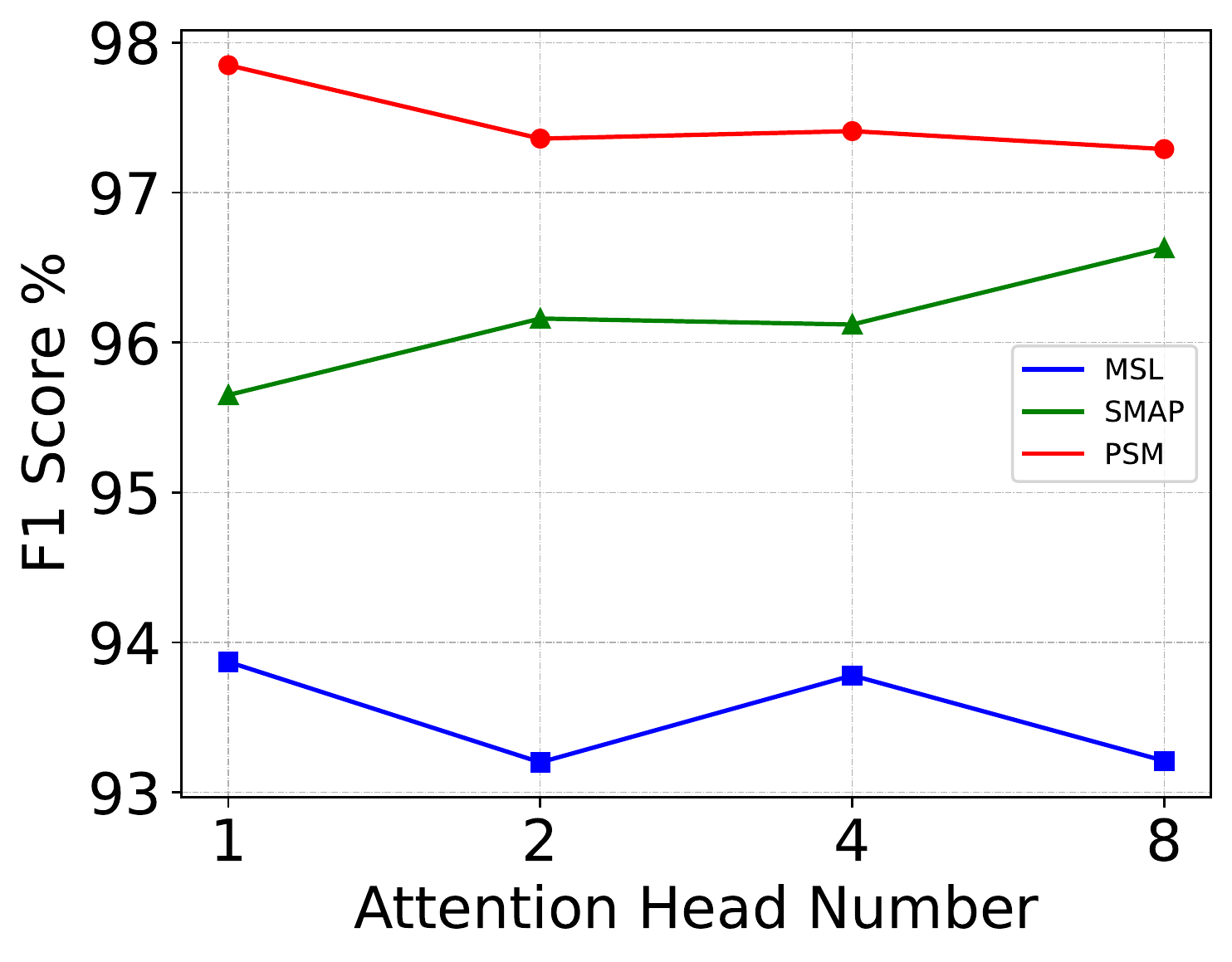}\label{fig:ab-head}}
	\subfigure[$d_{model}$ of attention]{\includegraphics[width=.195\textwidth]{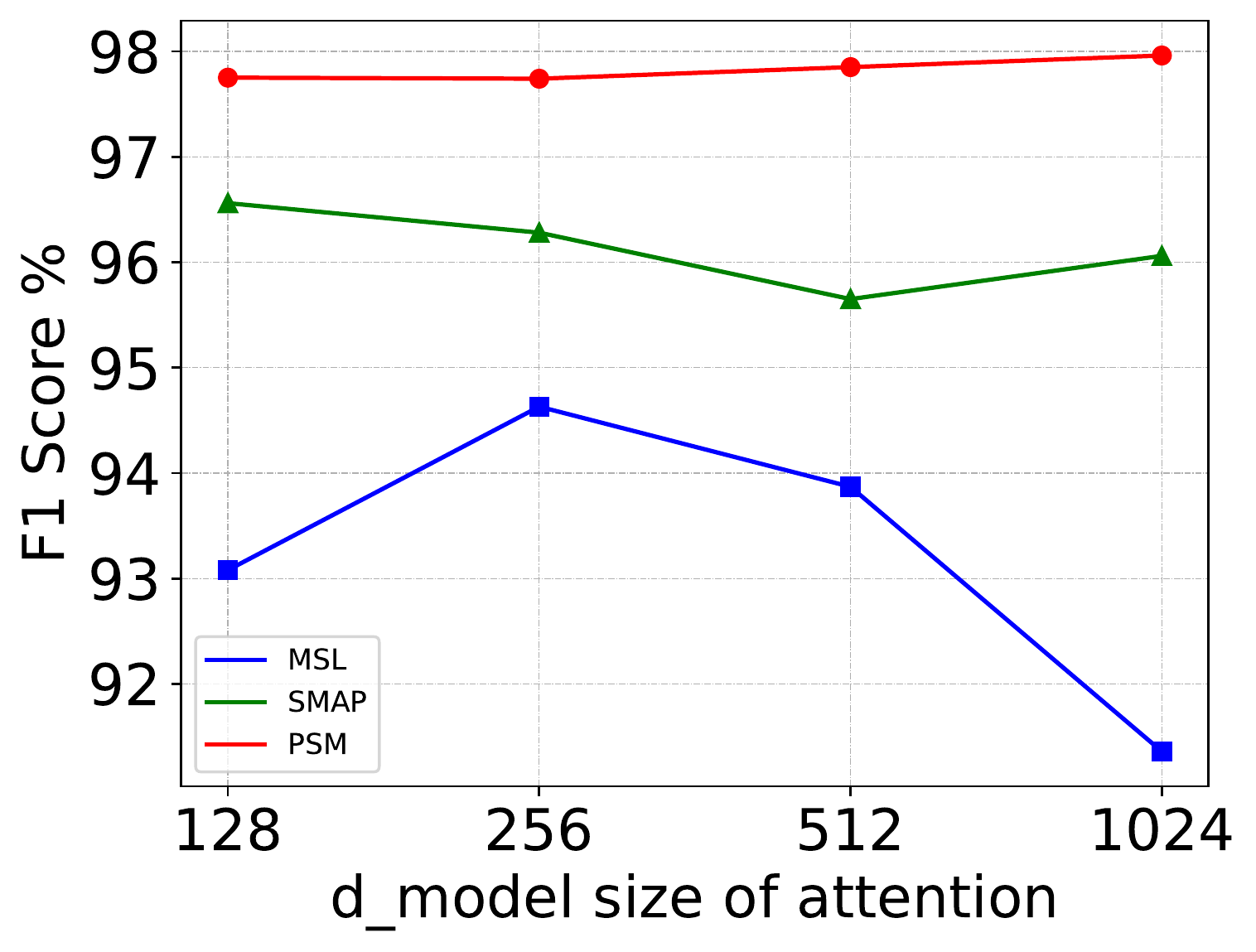}\label{fig:ab-atten}}
    \caption{Parameter sensitivity studies of main hyper-parameters in DCdetector.}
        \label{fig:ab} 
\end{figure*}
\begin{figure}[htbp]
	\centering
	\subfigure[Memory used]{\includegraphics[width=.225\textwidth]{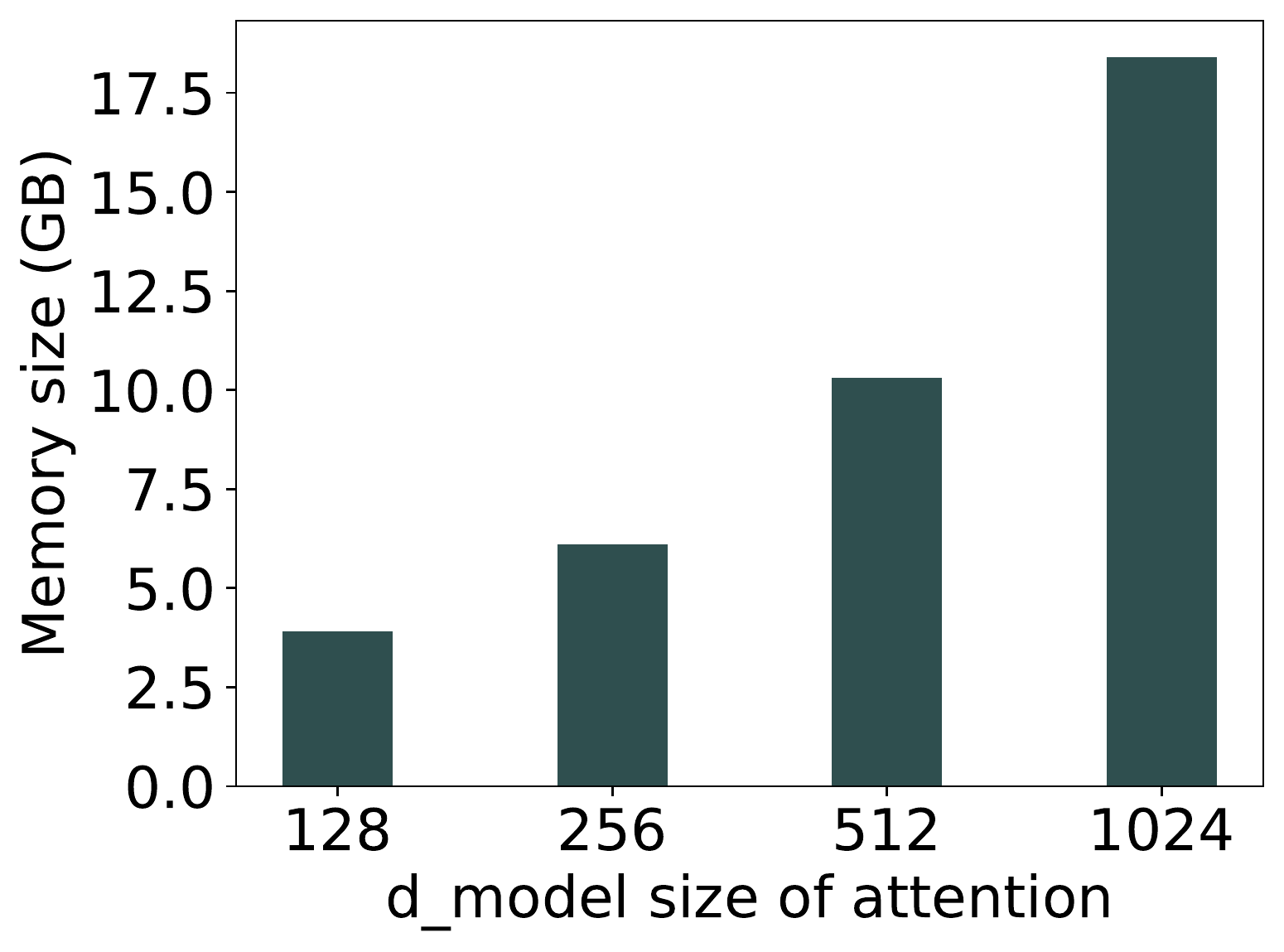}}\quad
	\subfigure[Time cost]{\includegraphics[width=.225\textwidth]{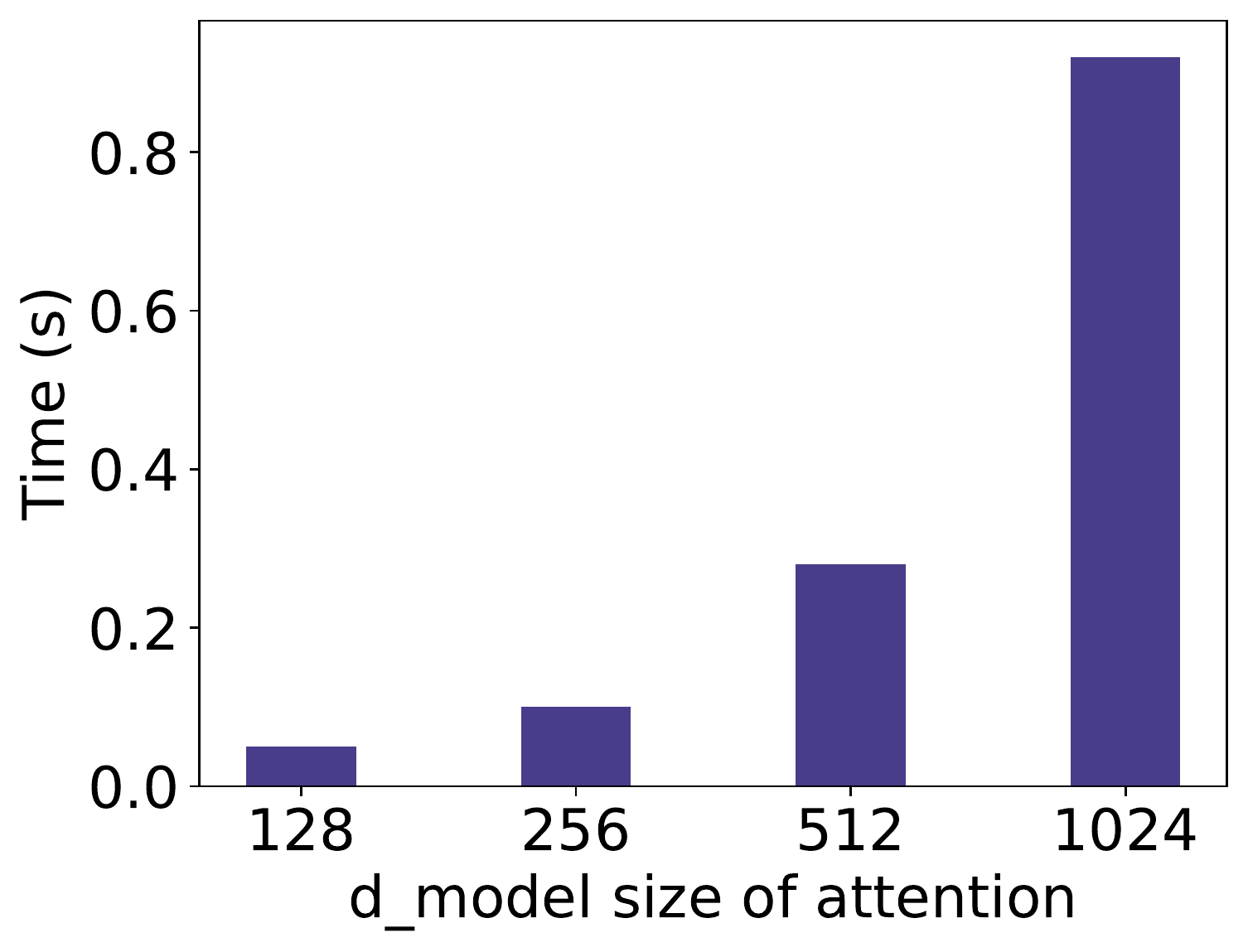}\label{fig:ab-time}}
        \caption{The averaged GPU memory cost and the averaged running time of 100 iterations during training with different $d_{model}$ sizes. } 
        \label{fig:mem-time}  
\end{figure}

\subsubsection{Ablation Studies}\label{sec:ablation}
Table~\ref{Tab: Ablation Stop Gradient} shows the ablation study of stop gradient. According to the loss function definition in Section~\ref{sec:rd}, we use two stop gradient modules in $\mathcal{L}\{\mathcal{P}, \mathcal{N}; \mathcal{X}\}$, noted as stop gradient in patch-wise branch and in-patch branch, respectively. With two-stop gradient modules, we can see that DCdetector gains the best performances. If no stop gradient is contained, DCdetector still works and does not fall into a trivial solution. Moreover, in such a setting, it outperforms all the baselines except Anomaly Transformer. 
Besides, we also conduct an ablation study on how the two main preprocessing methods (bilateral filter for denoising and instance normalization for normalization) affect the performance of our method in Table \ref{Tab: Ablation Preprocess results}. It can be seen that either of them slightly improves the performance of our model when used individually. However, if they are utilized simultaneously, the performance degrades. Therefore, our final DCdetector only contains the instance normalization module for preprocessing. More ablation studies on multi-scale patching, window size, attention head, embedding dimension, encoder layer, anomaly threshold, and metrics in loss function are left in Appendix~\ref{sec:appendix_extra_ablation}.

\subsubsection{Visual Analysis}

We show how DCdetector works by visualizing different anomalies in Figure~\ref{fig:case2}. We use the synthetic data generation methods reported in \cite{lai2021revisiting} to generate univariate time series with different types of anomalies, including point-wise anomalies (global point and contextual point anomalies) and pattern-wise anomalies (seasonal, group, and trend anomalies)~\cite{lai2021revisiting}. It can be seen that DCdetector can robustly detect various anomalies better from normal points with relatively higher anomaly scores.

\subsubsection{Parameter Sensitivity}
We also study the parameter sensitivity of the DCdetector. 
Figure~\ref{fig:ab-win} shows the performance under different window sizes. 
As discussed, a single point can not be taken as an instance in a time series. Window segmentation is widely used in the analysis, and window size is a significant parameter. 
For our primary evaluation, the window size is usually set as 60 or 100. Nevertheless, results in Figure~\ref{fig:ab-win} demonstrate that DCdetector is robust with a wide range of window sizes (from 30 to 210). Actually, in the window size range [45, 195], the performances fluctuate less than 2.3\%.
Figure~\ref{fig:ab-scale} shows the performance under different multi-scale sizes. 
Horizontal coordinate is the patch-size combination used in multi-scale attention which means we combine several dual-attention modules with a given patch-size combination. Unlike window size, the multi-scale design contributes to the final performance of the DCdetector, and different patch-size combinations lead to different performances. Note that when studying the parameter sensitivity of window size, the scale size is fixed as [3,5]. When studying the parameter sensitivity of scale size, the window size is fixed at 60. 
Figure~\ref{fig:ab-layer} shows the performance under different numbers of encoder layers, since many deep neural networks' performances are affected by the layer number. 
Figure~\ref{fig:ab-head} and Figure~\ref{fig:ab-atten} show model performances with different head numbers or $d_{model}$ sizes in attention. It can be seen that DCdetector achieves the best performance with a small attention head number and $d_{model}$ size. The memory and time usages with different $d_{model}$ sizes are shown in Figure~\ref{fig:mem-time}. Based on Figure~\ref{fig:mem-time} and Figure~\ref{fig:ab-atten}, we set the dimension of the hidden state $d_{model}=256$ for the performance-complexity trade-off, and it can be seen that DCdetector can work quite well under $d_{model}=256$ with efficient running time and small memory consumption.

\section{Conclusion}

This paper proposes a novel algorithm named DCdetector for time-series anomaly detection. We design a contrastive learning-based dual-branch attention structure in DCdetector to learn a permutation invariant representation. 
Such representation enlarges the differences between normal points and anomalies, improving detection accuracy. 
Besides, two additional designs: multiscale and channel independence patching, are implemented to enhance the performance. 
Moreover, we propose a pure contrastive loss function without reconstruction error, which empirically proves the effectiveness of contrastive representation compared to the widely used reconstructive one. 
Lastly, extensive experiments show that DCdetector achieves the best or comparable performance on eight benchmark datasets compared to various state-of-the-art algorithms. 

\section{Acknowledgements}
This work was supported by Alibaba Group through Alibaba Research Intern Program.

\bibliographystyle{ACM-Reference-Format}
\clearpage
\bibliography{8_reference}


\begin{thebibliography}{92}


\ifx \showCODEN    \undefined \def \showCODEN     #1{\unskip}     \fi
\ifx \showDOI      \undefined \def \showDOI       #1{#1}\fi
\ifx \showISBNx    \undefined \def \showISBNx     #1{\unskip}     \fi
\ifx \showISBNxiii \undefined \def \showISBNxiii  #1{\unskip}     \fi
\ifx \showISSN     \undefined \def \showISSN      #1{\unskip}     \fi
\ifx \showLCCN     \undefined \def \showLCCN      #1{\unskip}     \fi
\ifx \shownote     \undefined \def \shownote      #1{#1}          \fi
\ifx \showarticletitle \undefined \def \showarticletitle #1{#1}   \fi
\ifx \showURL      \undefined \def \showURL       {\relax}        \fi
\providecommand\bibfield[2]{#2}
\providecommand\bibinfo[2]{#2}
\providecommand\natexlab[1]{#1}
\providecommand\showeprint[2][]{arXiv:#2}

\bibitem[Abdulaal et~al\mbox{.}(2021)]%
        {abdulaal2021practical}
\bibfield{author}{\bibinfo{person}{Ahmed Abdulaal}, \bibinfo{person}{Zhuanghua
  Liu}, {and} \bibinfo{person}{Tomer Lancewicki}.}
  \bibinfo{year}{2021}\natexlab{}.
\newblock \showarticletitle{Practical approach to asynchronous multivariate
  time series anomaly detection and localization}. In
  \bibinfo{booktitle}{\emph{Proceedings of the 27th ACM SIGKDD Conference on
  Knowledge Discovery \& Data Mining}}. \bibinfo{pages}{2485--2494}.
\newblock


\bibitem[Adams and MacKay(2007)]%
        {adams2007bayesian}
\bibfield{author}{\bibinfo{person}{Ryan~Prescott Adams} {and}
  \bibinfo{person}{David~JC MacKay}.} \bibinfo{year}{2007}\natexlab{}.
\newblock \showarticletitle{Bayesian online changepoint detection}.
\newblock \bibinfo{journal}{\emph{arXiv preprint arXiv:0710.3742}}
  (\bibinfo{year}{2007}).
\newblock


\bibitem[Anandakrishnan et~al\mbox{.}(2018)]%
        {anandakrishnan2018anomaly}
\bibfield{author}{\bibinfo{person}{Archana Anandakrishnan},
  \bibinfo{person}{Senthil Kumar}, \bibinfo{person}{Alexander Statnikov},
  \bibinfo{person}{Tanveer Faruquie}, {and} \bibinfo{person}{Di Xu}.}
  \bibinfo{year}{2018}\natexlab{}.
\newblock \showarticletitle{Anomaly detection in finance: editors’
  introduction}. In \bibinfo{booktitle}{\emph{KDD 2017 Workshop on Anomaly
  Detection in Finance}}. PMLR, \bibinfo{pages}{1--7}.
\newblock


\bibitem[Anderson(1976)]%
        {anderson1976time}
\bibfield{author}{\bibinfo{person}{OD Anderson}.}
  \bibinfo{year}{1976}\natexlab{}.
\newblock \bibinfo{title}{Time-Series. 2nd edn.}
\newblock
\newblock


\bibitem[Angryk et~al\mbox{.}(2020)]%
        {DVN/EBCFKM_2020}
\bibfield{author}{\bibinfo{person}{Rafal Angryk}, \bibinfo{person}{Petrus
  Martens}, \bibinfo{person}{Berkay Aydin}, \bibinfo{person}{Dustin Kempton},
  \bibinfo{person}{Sushant Mahajan}, \bibinfo{person}{Sunitha Basodi},
  \bibinfo{person}{Azim Ahmadzadeh}, \bibinfo{person}{Xumin Cai},
  \bibinfo{person}{Soukaina Filali~Boubrahimi}, \bibinfo{person}{Shah~Muhammad
  Hamdi}, \bibinfo{person}{Micheal Schuh}, {and} \bibinfo{person}{Manolis
  Georgoulis}.} \bibinfo{year}{2020}\natexlab{}.
\newblock \bibinfo{title}{{SWAN-SF}}.
\newblock
\newblock
\urldef\tempurl%
\url{https://doi.org/10.7910/DVN/EBCFKM}
\showDOI{\tempurl}


\bibitem[Bl{\'a}zquez-Garc{\'\i}a et~al\mbox{.}(2021)]%
        {blazquez2021review}
\bibfield{author}{\bibinfo{person}{Ane Bl{\'a}zquez-Garc{\'\i}a},
  \bibinfo{person}{Angel Conde}, \bibinfo{person}{Usue Mori}, {and}
  \bibinfo{person}{Jose~A Lozano}.} \bibinfo{year}{2021}\natexlab{}.
\newblock \showarticletitle{A review on outlier/anomaly detection in time
  series data}.
\newblock \bibinfo{journal}{\emph{ACM Computing Surveys (CSUR)}}
  \bibinfo{volume}{54}, \bibinfo{number}{3} (\bibinfo{year}{2021}),
  \bibinfo{pages}{1--33}.
\newblock


\bibitem[Boniol and Palpanas(2020)]%
        {Boniol2020Series2GraphGS}
\bibfield{author}{\bibinfo{person}{Paul Boniol} {and} \bibinfo{person}{Themis
  Palpanas}.} \bibinfo{year}{2020}\natexlab{}.
\newblock \showarticletitle{Series2Graph: Graph-based Subsequence Anomaly
  Detection for Time Series}.
\newblock \bibinfo{journal}{\emph{ArXiv}}  \bibinfo{volume}{abs/2207.12208}
  (\bibinfo{year}{2020}).
\newblock


\bibitem[Bontemps et~al\mbox{.}(2016)]%
        {bontemps2016collective}
\bibfield{author}{\bibinfo{person}{Lo{\"\i}c Bontemps},
  \bibinfo{person}{Van~Loi Cao}, \bibinfo{person}{James McDermott}, {and}
  \bibinfo{person}{Nhien-An Le-Khac}.} \bibinfo{year}{2016}\natexlab{}.
\newblock \showarticletitle{Collective anomaly detection based on long
  short-term memory recurrent neural networks}. In
  \bibinfo{booktitle}{\emph{International conference on future data and
  security engineering}}. Springer, \bibinfo{pages}{141--152}.
\newblock


\bibitem[Box and Pierce(1970)]%
        {box1970distribution}
\bibfield{author}{\bibinfo{person}{George~EP Box} {and}
  \bibinfo{person}{David~A Pierce}.} \bibinfo{year}{1970}\natexlab{}.
\newblock \showarticletitle{Distribution of residual autocorrelations in
  autoregressive-integrated moving average time series models}.
\newblock \bibinfo{journal}{\emph{Journal of the American statistical
  Association}} \bibinfo{volume}{65}, \bibinfo{number}{332}
  (\bibinfo{year}{1970}), \bibinfo{pages}{1509--1526}.
\newblock


\bibitem[Breunig et~al\mbox{.}(2000)]%
        {breunig2000lof}
\bibfield{author}{\bibinfo{person}{Markus~M Breunig},
  \bibinfo{person}{Hans-Peter Kriegel}, \bibinfo{person}{Raymond~T Ng}, {and}
  \bibinfo{person}{J{\"o}rg Sander}.} \bibinfo{year}{2000}\natexlab{}.
\newblock \showarticletitle{LOF: identifying density-based local outliers}. In
  \bibinfo{booktitle}{\emph{Proceedings of the 2000 ACM SIGMOD international
  conference on Management of data}}. \bibinfo{pages}{93--104}.
\newblock


\bibitem[Campos et~al\mbox{.}(2021)]%
        {campos2021unsupervised}
\bibfield{author}{\bibinfo{person}{David Campos}, \bibinfo{person}{Tung Kieu},
  \bibinfo{person}{Chenjuan Guo}, \bibinfo{person}{Feiteng Huang},
  \bibinfo{person}{Kai Zheng}, \bibinfo{person}{Bin Yang}, {and}
  \bibinfo{person}{Christian~S Jensen}.} \bibinfo{year}{2021}\natexlab{}.
\newblock \showarticletitle{Unsupervised Time Series Outlier Detection with
  Diversity-Driven Convolutional Ensembles--Extended Version}.
\newblock \bibinfo{journal}{\emph{arXiv preprint arXiv:2111.11108}}
  (\bibinfo{year}{2021}).
\newblock


\bibitem[Canizo et~al\mbox{.}(2019)]%
        {canizo2019multi}
\bibfield{author}{\bibinfo{person}{Mikel Canizo}, \bibinfo{person}{Isaac
  Triguero}, \bibinfo{person}{Angel Conde}, {and} \bibinfo{person}{Enrique
  Onieva}.} \bibinfo{year}{2019}\natexlab{}.
\newblock \showarticletitle{Multi-head CNN--RNN for multi-time series anomaly
  detection: An industrial case study}.
\newblock \bibinfo{journal}{\emph{Neurocomputing}}  \bibinfo{volume}{363}
  (\bibinfo{year}{2019}), \bibinfo{pages}{246--260}.
\newblock


\bibitem[Caron et~al\mbox{.}(2020)]%
        {caron2020unsupervised}
\bibfield{author}{\bibinfo{person}{Mathilde Caron}, \bibinfo{person}{Ishan
  Misra}, \bibinfo{person}{Julien Mairal}, \bibinfo{person}{Priya Goyal},
  \bibinfo{person}{Piotr Bojanowski}, {and} \bibinfo{person}{Armand Joulin}.}
  \bibinfo{year}{2020}\natexlab{}.
\newblock \showarticletitle{Unsupervised learning of visual features by
  contrasting cluster assignments}.
\newblock \bibinfo{journal}{\emph{Advances in neural information processing
  systems}}  \bibinfo{volume}{33} (\bibinfo{year}{2020}),
  \bibinfo{pages}{9912--9924}.
\newblock


\bibitem[Chauhan and Vig(2015)]%
        {chauhan2015anomaly}
\bibfield{author}{\bibinfo{person}{Sucheta Chauhan} {and}
  \bibinfo{person}{Lovekesh Vig}.} \bibinfo{year}{2015}\natexlab{}.
\newblock \showarticletitle{Anomaly detection in ECG time signals via deep long
  short-term memory networks}. In \bibinfo{booktitle}{\emph{2015 IEEE
  international conference on data science and advanced analytics (DSAA)}}.
  IEEE, \bibinfo{pages}{1--7}.
\newblock


\bibitem[Chen et~al\mbox{.}(2020)]%
        {chen2020simple}
\bibfield{author}{\bibinfo{person}{Ting Chen}, \bibinfo{person}{Simon
  Kornblith}, \bibinfo{person}{Mohammad Norouzi}, {and}
  \bibinfo{person}{Geoffrey Hinton}.} \bibinfo{year}{2020}\natexlab{}.
\newblock \showarticletitle{A simple framework for contrastive learning of
  visual representations}. In \bibinfo{booktitle}{\emph{International
  conference on machine learning}}. PMLR, \bibinfo{pages}{1597--1607}.
\newblock


\bibitem[Chen et~al\mbox{.}(2021b)]%
        {chen2021daemon}
\bibfield{author}{\bibinfo{person}{Xuanhao Chen}, \bibinfo{person}{Liwei Deng},
  \bibinfo{person}{Feiteng Huang}, \bibinfo{person}{Chengwei Zhang},
  \bibinfo{person}{Zongquan Zhang}, \bibinfo{person}{Yan Zhao}, {and}
  \bibinfo{person}{Kai Zheng}.} \bibinfo{year}{2021}\natexlab{b}.
\newblock \showarticletitle{Daemon: Unsupervised anomaly detection and
  interpretation for multivariate time series}. In
  \bibinfo{booktitle}{\emph{2021 IEEE 37th International Conference on Data
  Engineering (ICDE)}}. IEEE, \bibinfo{pages}{2225--2230}.
\newblock


\bibitem[Chen and He(2021)]%
        {chen2021exploring}
\bibfield{author}{\bibinfo{person}{Xinlei Chen} {and} \bibinfo{person}{Kaiming
  He}.} \bibinfo{year}{2021}\natexlab{}.
\newblock \showarticletitle{Exploring simple siamese representation learning}.
  In \bibinfo{booktitle}{\emph{Proceedings of the IEEE/CVF Conference on
  Computer Vision and Pattern Recognition}}. \bibinfo{pages}{15750--15758}.
\newblock


\bibitem[Chen et~al\mbox{.}(2021a)]%
        {chen2021learning}
\bibfield{author}{\bibinfo{person}{Zekai Chen}, \bibinfo{person}{Dingshuo
  Chen}, \bibinfo{person}{Xiao Zhang}, \bibinfo{person}{Zixuan Yuan}, {and}
  \bibinfo{person}{Xiuzhen Cheng}.} \bibinfo{year}{2021}\natexlab{a}.
\newblock \showarticletitle{Learning graph structures with transformer for
  multivariate time-series anomaly detection in IoT}.
\newblock \bibinfo{journal}{\emph{IEEE Internet of Things Journal}}
  \bibinfo{volume}{9}, \bibinfo{number}{12} (\bibinfo{year}{2021}),
  \bibinfo{pages}{9179--9189}.
\newblock


\bibitem[Cheng et~al\mbox{.}(2009)]%
        {cheng2009detection}
\bibfield{author}{\bibinfo{person}{Haibin Cheng}, \bibinfo{person}{Pang-Ning
  Tan}, \bibinfo{person}{Christopher Potter}, {and} \bibinfo{person}{Steven
  Klooster}.} \bibinfo{year}{2009}\natexlab{}.
\newblock \showarticletitle{Detection and characterization of anomalies in
  multivariate time series}. In \bibinfo{booktitle}{\emph{Proceedings of the
  2009 SIAM international conference on data mining}}. SIAM,
  \bibinfo{pages}{413--424}.
\newblock


\bibitem[Cook et~al\mbox{.}(2019)]%
        {cook2019anomaly}
\bibfield{author}{\bibinfo{person}{Andrew~A Cook}, \bibinfo{person}{G{\"o}ksel
  M{\i}s{\i}rl{\i}}, {and} \bibinfo{person}{Zhong Fan}.}
  \bibinfo{year}{2019}\natexlab{}.
\newblock \showarticletitle{Anomaly detection for IoT time-series data: A
  survey}.
\newblock \bibinfo{journal}{\emph{IEEE Internet of Things Journal}}
  \bibinfo{volume}{7}, \bibinfo{number}{7} (\bibinfo{year}{2019}),
  \bibinfo{pages}{6481--6494}.
\newblock


\bibitem[Dau et~al\mbox{.}(2019)]%
        {dau2019ucr}
\bibfield{author}{\bibinfo{person}{Hoang~Anh Dau}, \bibinfo{person}{Anthony
  Bagnall}, \bibinfo{person}{Kaveh Kamgar}, \bibinfo{person}{Chin-Chia~Michael
  Yeh}, \bibinfo{person}{Yan Zhu}, \bibinfo{person}{Shaghayegh Gharghabi},
  \bibinfo{person}{Chotirat~Ann Ratanamahatana}, {and} \bibinfo{person}{Eamonn
  Keogh}.} \bibinfo{year}{2019}\natexlab{}.
\newblock \showarticletitle{The UCR time series archive}.
\newblock \bibinfo{journal}{\emph{IEEE/CAA Journal of Automatica Sinica}}
  \bibinfo{volume}{6}, \bibinfo{number}{6} (\bibinfo{year}{2019}),
  \bibinfo{pages}{1293--1305}.
\newblock


\bibitem[Deldari et~al\mbox{.}(2021)]%
        {deldari2021time}
\bibfield{author}{\bibinfo{person}{Shohreh Deldari}, \bibinfo{person}{Daniel~V
  Smith}, \bibinfo{person}{Hao Xue}, {and} \bibinfo{person}{Flora~D Salim}.}
  \bibinfo{year}{2021}\natexlab{}.
\newblock \showarticletitle{Time series change point detection with
  self-supervised contrastive predictive coding}. In
  \bibinfo{booktitle}{\emph{Proceedings of the Web Conference 2021}}.
  \bibinfo{pages}{3124--3135}.
\newblock


\bibitem[Deng and Hooi(2021a)]%
        {Deng2021GraphNN}
\bibfield{author}{\bibinfo{person}{Ailin Deng} {and} \bibinfo{person}{Bryan
  Hooi}.} \bibinfo{year}{2021}\natexlab{a}.
\newblock \showarticletitle{Graph Neural Network-Based Anomaly Detection in
  Multivariate Time Series}. In \bibinfo{booktitle}{\emph{AAAI Conference on
  Artificial Intelligence}}.
\newblock


\bibitem[Deng and Hooi(2021b)]%
        {deng2021graph}
\bibfield{author}{\bibinfo{person}{Ailin Deng} {and} \bibinfo{person}{Bryan
  Hooi}.} \bibinfo{year}{2021}\natexlab{b}.
\newblock \showarticletitle{Graph neural network-based anomaly detection in
  multivariate time series}. In \bibinfo{booktitle}{\emph{Proceedings of the
  AAAI conference on artificial intelligence}}, Vol.~\bibinfo{volume}{35}.
  \bibinfo{pages}{4027--4035}.
\newblock


\bibitem[Elsayed et~al\mbox{.}(2021)]%
        {elsayed2021we}
\bibfield{author}{\bibinfo{person}{Shereen Elsayed}, \bibinfo{person}{Daniela
  Thyssens}, \bibinfo{person}{Ahmed Rashed}, \bibinfo{person}{Hadi~Samer
  Jomaa}, {and} \bibinfo{person}{Lars Schmidt-Thieme}.}
  \bibinfo{year}{2021}\natexlab{}.
\newblock \showarticletitle{Do we really need deep learning models for time
  series forecasting?}
\newblock \bibinfo{journal}{\emph{arXiv preprint arXiv:2101.02118}}
  (\bibinfo{year}{2021}).
\newblock


\bibitem[Gao et~al\mbox{.}(2020)]%
        {jingkun20_TAD}
\bibfield{author}{\bibinfo{person}{Jingkun Gao}, \bibinfo{person}{Xiaomin
  Song}, \bibinfo{person}{Qingsong Wen}, \bibinfo{person}{Pichao Wang},
  \bibinfo{person}{Liang Sun}, {and} \bibinfo{person}{Huan Xu}.}
  \bibinfo{year}{2020}\natexlab{}.
\newblock \showarticletitle{{RobustTAD}: Robust time series anomaly detection
  via decomposition and convolutional neural networks}.
\newblock \bibinfo{journal}{\emph{KDD Workshop MileTS}} (\bibinfo{year}{2020}).
\newblock


\bibitem[Golmohammadi and Zaiane(2015)]%
        {golmohammadi2015time}
\bibfield{author}{\bibinfo{person}{Koosha Golmohammadi} {and}
  \bibinfo{person}{Osmar~R Zaiane}.} \bibinfo{year}{2015}\natexlab{}.
\newblock \showarticletitle{Time series contextual anomaly detection for
  detecting market manipulation in stock market}. In
  \bibinfo{booktitle}{\emph{2015 IEEE international conference on data science
  and advanced analytics (DSAA)}}. IEEE, \bibinfo{pages}{1--10}.
\newblock


\bibitem[Grill et~al\mbox{.}(2020)]%
        {grill2020bootstrap}
\bibfield{author}{\bibinfo{person}{Jean-Bastien Grill},
  \bibinfo{person}{Florian Strub}, \bibinfo{person}{Florent Altch{\'e}},
  \bibinfo{person}{Corentin Tallec}, \bibinfo{person}{Pierre Richemond},
  \bibinfo{person}{Elena Buchatskaya}, \bibinfo{person}{Carl Doersch},
  \bibinfo{person}{Bernardo Avila~Pires}, \bibinfo{person}{Zhaohan Guo},
  \bibinfo{person}{Mohammad Gheshlaghi~Azar}, {et~al\mbox{.}}}
  \bibinfo{year}{2020}\natexlab{}.
\newblock \showarticletitle{Bootstrap your own latent-a new approach to
  self-supervised learning}.
\newblock \bibinfo{journal}{\emph{Advances in neural information processing
  systems}}  \bibinfo{volume}{33} (\bibinfo{year}{2020}),
  \bibinfo{pages}{21271--21284}.
\newblock


\bibitem[He et~al\mbox{.}(2020)]%
        {he2020momentum}
\bibfield{author}{\bibinfo{person}{Kaiming He}, \bibinfo{person}{Haoqi Fan},
  \bibinfo{person}{Yuxin Wu}, \bibinfo{person}{Saining Xie}, {and}
  \bibinfo{person}{Ross Girshick}.} \bibinfo{year}{2020}\natexlab{}.
\newblock \showarticletitle{Momentum contrast for unsupervised visual
  representation learning}. In \bibinfo{booktitle}{\emph{Proceedings of the
  IEEE/CVF conference on computer vision and pattern recognition}}.
  \bibinfo{pages}{9729--9738}.
\newblock


\bibitem[Huang et~al\mbox{.}(2018)]%
        {huang2018towards}
\bibfield{author}{\bibinfo{person}{Chengqiang Huang}, \bibinfo{person}{Yulei
  Wu}, \bibinfo{person}{Yuan Zuo}, \bibinfo{person}{Ke Pei}, {and}
  \bibinfo{person}{Geyong Min}.} \bibinfo{year}{2018}\natexlab{}.
\newblock \showarticletitle{Towards experienced anomaly detector through
  reinforcement learning}. In \bibinfo{booktitle}{\emph{Proceedings of the AAAI
  Conference on Artificial Intelligence}}, Vol.~\bibinfo{volume}{32}.
\newblock


\bibitem[Huet et~al\mbox{.}(2022)]%
        {huet2022local}
\bibfield{author}{\bibinfo{person}{Alexis Huet}, \bibinfo{person}{Jose~Manuel
  Navarro}, {and} \bibinfo{person}{Dario Rossi}.}
  \bibinfo{year}{2022}\natexlab{}.
\newblock \showarticletitle{Local Evaluation of Time Series Anomaly Detection
  Algorithms}. In \bibinfo{booktitle}{\emph{Proceedings of the 28th ACM SIGKDD
  Conference on Knowledge Discovery and Data Mining}}.
  \bibinfo{pages}{635--645}.
\newblock


\bibitem[Hundman et~al\mbox{.}(2018)]%
        {hundman2018detecting}
\bibfield{author}{\bibinfo{person}{Kyle Hundman}, \bibinfo{person}{Valentino
  Constantinou}, \bibinfo{person}{Christopher Laporte}, \bibinfo{person}{Ian
  Colwell}, {and} \bibinfo{person}{Tom Soderstrom}.}
  \bibinfo{year}{2018}\natexlab{}.
\newblock \showarticletitle{Detecting spacecraft anomalies using lstms and
  nonparametric dynamic thresholding}. In \bibinfo{booktitle}{\emph{Proceedings
  of the 24th ACM SIGKDD international conference on knowledge discovery \&
  data mining}}. \bibinfo{pages}{387--395}.
\newblock


\bibitem[Jiao et~al\mbox{.}(2022)]%
        {jiao2022timeautoad}
\bibfield{author}{\bibinfo{person}{Yang Jiao}, \bibinfo{person}{Kai Yang},
  \bibinfo{person}{Dongjing Song}, {and} \bibinfo{person}{Dacheng Tao}.}
  \bibinfo{year}{2022}\natexlab{}.
\newblock \showarticletitle{Timeautoad: Autonomous anomaly detection with
  self-supervised contrastive loss for multivariate time series}.
\newblock \bibinfo{journal}{\emph{IEEE Transactions on Network Science and
  Engineering}} \bibinfo{volume}{9}, \bibinfo{number}{3}
  (\bibinfo{year}{2022}), \bibinfo{pages}{1604--1619}.
\newblock


\bibitem[Kant and Mahajan(2019)]%
        {kant2019time}
\bibfield{author}{\bibinfo{person}{Neha Kant} {and} \bibinfo{person}{Manish
  Mahajan}.} \bibinfo{year}{2019}\natexlab{}.
\newblock \showarticletitle{Time-series outlier detection using enhanced
  k-means in combination with pso algorithm}. In
  \bibinfo{booktitle}{\emph{Engineering Vibration, Communication and
  Information Processing: ICoEVCI 2018, India}}. Springer,
  \bibinfo{pages}{363--373}.
\newblock


\bibitem[Karczmarek et~al\mbox{.}(2020)]%
        {karczmarek2020k}
\bibfield{author}{\bibinfo{person}{Pawe{\l} Karczmarek}, \bibinfo{person}{Adam
  Kiersztyn}, \bibinfo{person}{Witold Pedrycz}, {and} \bibinfo{person}{Ebru
  Al}.} \bibinfo{year}{2020}\natexlab{}.
\newblock \showarticletitle{K-Means-based isolation forest}.
\newblock \bibinfo{journal}{\emph{Knowledge-based systems}}
  \bibinfo{volume}{195} (\bibinfo{year}{2020}), \bibinfo{pages}{105659}.
\newblock


\bibitem[Keogh et~al\mbox{.}(2021)]%
        {keogh2021multi}
\bibfield{author}{\bibinfo{person}{Eamonn Keogh}, \bibinfo{person}{Dutta~Roy
  Taposh}, \bibinfo{person}{U Naik}, {and} \bibinfo{person}{A Agrawal}.}
  \bibinfo{year}{2021}\natexlab{}.
\newblock \showarticletitle{Multi-dataset Time-Series Anomaly Detection
  Competition}. In \bibinfo{booktitle}{\emph{ACM SIGKDD International
  Conference on Knowledge Discovery and Data Mining. https://compete.
  hexagonml. com/practice/competition/39}}.
\newblock


\bibitem[Khan et~al\mbox{.}(2022)]%
        {khan2022contrastive}
\bibfield{author}{\bibinfo{person}{Adnan Khan}, \bibinfo{person}{Sarah
  AlBarri}, {and} \bibinfo{person}{Muhammad~Arslan Manzoor}.}
  \bibinfo{year}{2022}\natexlab{}.
\newblock \showarticletitle{Contrastive self-supervised learning: a survey on
  different architectures}. In \bibinfo{booktitle}{\emph{2022 2nd International
  Conference on Artificial Intelligence (ICAI)}}. IEEE, \bibinfo{pages}{1--6}.
\newblock


\bibitem[Kim et~al\mbox{.}(2021)]%
        {kim2021reversible}
\bibfield{author}{\bibinfo{person}{Taesung Kim}, \bibinfo{person}{Jinhee Kim},
  \bibinfo{person}{Yunwon Tae}, \bibinfo{person}{Cheonbok Park},
  \bibinfo{person}{Jang-Ho Choi}, {and} \bibinfo{person}{Jaegul Choo}.}
  \bibinfo{year}{2021}\natexlab{}.
\newblock \showarticletitle{Reversible instance normalization for accurate
  time-series forecasting against distribution shift}. In
  \bibinfo{booktitle}{\emph{International Conference on Learning
  Representations}}.
\newblock


\bibitem[Kingma and Ba(2014)]%
        {kingma2014adam}
\bibfield{author}{\bibinfo{person}{Diederik~P Kingma} {and}
  \bibinfo{person}{Jimmy Ba}.} \bibinfo{year}{2014}\natexlab{}.
\newblock \showarticletitle{Adam: A method for stochastic optimization}.
\newblock \bibinfo{journal}{\emph{arXiv preprint arXiv:1412.6980}}
  (\bibinfo{year}{2014}).
\newblock


\bibitem[Lai et~al\mbox{.}(2021)]%
        {lai2021revisiting}
\bibfield{author}{\bibinfo{person}{Kwei-Herng Lai}, \bibinfo{person}{Daochen
  Zha}, \bibinfo{person}{Junjie Xu}, \bibinfo{person}{Yue Zhao},
  \bibinfo{person}{Guanchu Wang}, {and} \bibinfo{person}{Xia Hu}.}
  \bibinfo{year}{2021}\natexlab{}.
\newblock \showarticletitle{Revisiting time series outlier detection:
  Definitions and benchmarks}. In \bibinfo{booktitle}{\emph{Thirty-fifth
  Conference on Neural Information Processing Systems Datasets and Benchmarks
  Track (Round 1)}}.
\newblock


\bibitem[Li et~al\mbox{.}(2019a)]%
        {li2019mad}
\bibfield{author}{\bibinfo{person}{Dan Li}, \bibinfo{person}{Dacheng Chen},
  \bibinfo{person}{Baihong Jin}, \bibinfo{person}{Lei Shi},
  \bibinfo{person}{Jonathan Goh}, {and} \bibinfo{person}{See-Kiong Ng}.}
  \bibinfo{year}{2019}\natexlab{a}.
\newblock \showarticletitle{MAD-GAN: Multivariate anomaly detection for time
  series data with generative adversarial networks}. In
  \bibinfo{booktitle}{\emph{Artificial Neural Networks and Machine
  Learning--ICANN 2019: Text and Time Series: 28th International Conference on
  Artificial Neural Networks, Munich, Germany, September 17--19, 2019,
  Proceedings, Part IV}}. Springer, \bibinfo{pages}{703--716}.
\newblock


\bibitem[Li et~al\mbox{.}(2022)]%
        {li2022learning}
\bibfield{author}{\bibinfo{person}{Longyuan Li}, \bibinfo{person}{Junchi Yan},
  \bibinfo{person}{Qingsong Wen}, \bibinfo{person}{Yaohui Jin}, {and}
  \bibinfo{person}{Xiaokang Yang}.} \bibinfo{year}{2022}\natexlab{}.
\newblock \showarticletitle{Learning robust deep state space for unsupervised
  anomaly detection in contaminated time-series}.
\newblock \bibinfo{journal}{\emph{IEEE Transactions on Knowledge and Data
  Engineering}} (\bibinfo{year}{2022}).
\newblock


\bibitem[Li et~al\mbox{.}(2019b)]%
        {li2019enhancing}
\bibfield{author}{\bibinfo{person}{Shiyang Li}, \bibinfo{person}{Xiaoyong Jin},
  \bibinfo{person}{Yao Xuan}, \bibinfo{person}{Xiyou Zhou},
  \bibinfo{person}{Wenhu Chen}, \bibinfo{person}{Yu-Xiang Wang}, {and}
  \bibinfo{person}{Xifeng Yan}.} \bibinfo{year}{2019}\natexlab{b}.
\newblock \showarticletitle{Enhancing the locality and breaking the memory
  bottleneck of transformer on time series forecasting}.
\newblock \bibinfo{journal}{\emph{Advances in neural information processing
  systems}}  \bibinfo{volume}{32} (\bibinfo{year}{2019}).
\newblock


\bibitem[Li et~al\mbox{.}(2021a)]%
        {li2021block}
\bibfield{author}{\bibinfo{person}{Xing Li}, \bibinfo{person}{Qiquan Shi},
  \bibinfo{person}{Gang Hu}, \bibinfo{person}{Lei Chen}, \bibinfo{person}{Hui
  Mao}, \bibinfo{person}{Yiyuan Yang}, \bibinfo{person}{Mingxuan Yuan},
  \bibinfo{person}{Jia Zeng}, {and} \bibinfo{person}{Zhuo Cheng}.}
  \bibinfo{year}{2021}\natexlab{a}.
\newblock \showarticletitle{Block Access Pattern Discovery via Compressed Full
  Tensor Transformer}. In \bibinfo{booktitle}{\emph{Proceedings of the 30th ACM
  International Conference on Information \& Knowledge Management}}.
  \bibinfo{pages}{957--966}.
\newblock


\bibitem[Li et~al\mbox{.}(2021b)]%
        {li2021multivariate}
\bibfield{author}{\bibinfo{person}{Zhihan Li}, \bibinfo{person}{Youjian Zhao},
  \bibinfo{person}{Jiaqi Han}, \bibinfo{person}{Ya Su}, \bibinfo{person}{Rui
  Jiao}, \bibinfo{person}{Xidao Wen}, {and} \bibinfo{person}{Dan Pei}.}
  \bibinfo{year}{2021}\natexlab{b}.
\newblock \showarticletitle{Multivariate time series anomaly detection and
  interpretation using hierarchical inter-metric and temporal embedding}. In
  \bibinfo{booktitle}{\emph{Proceedings of the 27th ACM SIGKDD Conference on
  Knowledge Discovery \& Data Mining}}. \bibinfo{pages}{3220--3230}.
\newblock


\bibitem[Liu et~al\mbox{.}(2008)]%
        {liu2008isolation}
\bibfield{author}{\bibinfo{person}{Fei~Tony Liu}, \bibinfo{person}{Kai~Ming
  Ting}, {and} \bibinfo{person}{Zhi-Hua Zhou}.}
  \bibinfo{year}{2008}\natexlab{}.
\newblock \showarticletitle{Isolation forest}. In
  \bibinfo{booktitle}{\emph{2008 eighth ieee international conference on data
  mining}}. IEEE, \bibinfo{pages}{413--422}.
\newblock


\bibitem[Mathur and Tippenhauer(2016)]%
        {mathur2016swat}
\bibfield{author}{\bibinfo{person}{Aditya~P Mathur} {and}
  \bibinfo{person}{Nils~Ole Tippenhauer}.} \bibinfo{year}{2016}\natexlab{}.
\newblock \showarticletitle{SWaT: A water treatment testbed for research and
  training on ICS security}. In \bibinfo{booktitle}{\emph{2016 international
  workshop on cyber-physical systems for smart water networks (CySWater)}}.
  IEEE, \bibinfo{pages}{31--36}.
\newblock


\bibitem[Moritz et~al\mbox{.}(2018)]%
        {moritz2018gecco}
\bibfield{author}{\bibinfo{person}{Steffen Moritz}, \bibinfo{person}{Frederik
  Rehbach}, \bibinfo{person}{Sowmya Chandrasekaran}, \bibinfo{person}{Margarita
  Rebolledo}, {and} \bibinfo{person}{Thomas Bartz-Beielstein}.}
  \bibinfo{year}{2018}\natexlab{}.
\newblock \bibinfo{title}{GECCO Industrial Challenge 2018 Dataset: a water
  quality dataset for the “Internet of Things: Online Anomaly Detection for
  Drinking Water Quality” competition at the Genetic and Evolutionary
  Computation Conference 2018, Kyoto, Japan (2018)}.
\newblock
\newblock


\bibitem[Munir et~al\mbox{.}(2018)]%
        {munir2018deepant}
\bibfield{author}{\bibinfo{person}{Mohsin Munir}, \bibinfo{person}{Shoaib~Ahmed
  Siddiqui}, \bibinfo{person}{Andreas Dengel}, {and} \bibinfo{person}{Sheraz
  Ahmed}.} \bibinfo{year}{2018}\natexlab{}.
\newblock \showarticletitle{DeepAnT: A deep learning approach for unsupervised
  anomaly detection in time series}.
\newblock \bibinfo{journal}{\emph{Ieee Access}}  \bibinfo{volume}{7}
  (\bibinfo{year}{2018}), \bibinfo{pages}{1991--2005}.
\newblock


\bibitem[Nie et~al\mbox{.}(2022)]%
        {nie2022time}
\bibfield{author}{\bibinfo{person}{Yuqi Nie}, \bibinfo{person}{Nam~H Nguyen},
  \bibinfo{person}{Phanwadee Sinthong}, {and} \bibinfo{person}{Jayant
  Kalagnanam}.} \bibinfo{year}{2022}\natexlab{}.
\newblock \showarticletitle{A Time Series is Worth 64 Words: Long-term
  Forecasting with Transformers}.
\newblock \bibinfo{journal}{\emph{arXiv preprint arXiv:2211.14730}}
  (\bibinfo{year}{2022}).
\newblock


\bibitem[Niu et~al\mbox{.}(2020)]%
        {niu2020lstm}
\bibfield{author}{\bibinfo{person}{Zijian Niu}, \bibinfo{person}{Ke Yu}, {and}
  \bibinfo{person}{Xiaofei Wu}.} \bibinfo{year}{2020}\natexlab{}.
\newblock \showarticletitle{LSTM-based VAE-GAN for time-series anomaly
  detection}.
\newblock \bibinfo{journal}{\emph{Sensors}} \bibinfo{volume}{20},
  \bibinfo{number}{13} (\bibinfo{year}{2020}), \bibinfo{pages}{3738}.
\newblock


\bibitem[Paparrizos et~al\mbox{.}(2022)]%
        {paparrizos2022volume}
\bibfield{author}{\bibinfo{person}{John Paparrizos}, \bibinfo{person}{Paul
  Boniol}, \bibinfo{person}{Themis Palpanas}, \bibinfo{person}{Ruey~S Tsay},
  \bibinfo{person}{Aaron Elmore}, {and} \bibinfo{person}{Michael~J Franklin}.}
  \bibinfo{year}{2022}\natexlab{}.
\newblock \showarticletitle{Volume under the surface: a new accuracy evaluation
  measure for time-series anomaly detection}.
\newblock \bibinfo{journal}{\emph{Proceedings of the VLDB Endowment}}
  \bibinfo{volume}{15}, \bibinfo{number}{11} (\bibinfo{year}{2022}),
  \bibinfo{pages}{2774--2787}.
\newblock


\bibitem[Park et~al\mbox{.}(2018)]%
        {park2018multimodal}
\bibfield{author}{\bibinfo{person}{Daehyung Park}, \bibinfo{person}{Yuuna
  Hoshi}, {and} \bibinfo{person}{Charles~C Kemp}.}
  \bibinfo{year}{2018}\natexlab{}.
\newblock \showarticletitle{A multimodal anomaly detector for robot-assisted
  feeding using an lstm-based variational autoencoder}.
\newblock \bibinfo{journal}{\emph{IEEE Robotics and Automation Letters}}
  \bibinfo{volume}{3}, \bibinfo{number}{3} (\bibinfo{year}{2018}),
  \bibinfo{pages}{1544--1551}.
\newblock


\bibitem[Paszke et~al\mbox{.}(2019)]%
        {paszke2019pytorch}
\bibfield{author}{\bibinfo{person}{Adam Paszke}, \bibinfo{person}{Sam Gross},
  \bibinfo{person}{Francisco Massa}, \bibinfo{person}{Adam Lerer},
  \bibinfo{person}{James Bradbury}, \bibinfo{person}{Gregory Chanan},
  \bibinfo{person}{Trevor Killeen}, \bibinfo{person}{Zeming Lin},
  \bibinfo{person}{Natalia Gimelshein}, \bibinfo{person}{Luca Antiga},
  {et~al\mbox{.}}} \bibinfo{year}{2019}\natexlab{}.
\newblock \showarticletitle{Pytorch: An imperative style, high-performance deep
  learning library}.
\newblock \bibinfo{journal}{\emph{Advances in neural information processing
  systems}}  \bibinfo{volume}{32} (\bibinfo{year}{2019}).
\newblock


\bibitem[Perslev et~al\mbox{.}(2019)]%
        {perslev2019u}
\bibfield{author}{\bibinfo{person}{Mathias Perslev}, \bibinfo{person}{Michael
  Jensen}, \bibinfo{person}{Sune Darkner}, \bibinfo{person}{Poul~J{\o}rgen
  Jennum}, {and} \bibinfo{person}{Christian Igel}.}
  \bibinfo{year}{2019}\natexlab{}.
\newblock \showarticletitle{U-time: A fully convolutional network for time
  series segmentation applied to sleep staging}.
\newblock \bibinfo{journal}{\emph{Advances in Neural Information Processing
  Systems}}  \bibinfo{volume}{32} (\bibinfo{year}{2019}).
\newblock


\bibitem[Phillips and Jin(2015)]%
        {phillips2015business}
\bibfield{author}{\bibinfo{person}{Peter~CB Phillips} {and}
  \bibinfo{person}{Sainan Jin}.} \bibinfo{year}{2015}\natexlab{}.
\newblock \showarticletitle{Business cycles, trend elimination, and the HP
  filter}.
\newblock  (\bibinfo{year}{2015}).
\newblock


\bibitem[Ren et~al\mbox{.}(2019)]%
        {ren2019time}
\bibfield{author}{\bibinfo{person}{Hansheng Ren}, \bibinfo{person}{Bixiong Xu},
  \bibinfo{person}{Yujing Wang}, \bibinfo{person}{Chao Yi},
  \bibinfo{person}{Congrui Huang}, \bibinfo{person}{Xiaoyu Kou},
  \bibinfo{person}{Tony Xing}, \bibinfo{person}{Mao Yang}, \bibinfo{person}{Jie
  Tong}, {and} \bibinfo{person}{Qi Zhang}.} \bibinfo{year}{2019}\natexlab{}.
\newblock \showarticletitle{Time-series anomaly detection service at
  microsoft}. In \bibinfo{booktitle}{\emph{Proceedings of the 25th ACM SIGKDD
  international conference on knowledge discovery \& data mining}}.
  \bibinfo{pages}{3009--3017}.
\newblock


\bibitem[Rousseeuw and Leroy(2005)]%
        {rousseeuw2005robust}
\bibfield{author}{\bibinfo{person}{Peter~J Rousseeuw} {and}
  \bibinfo{person}{Annick~M Leroy}.} \bibinfo{year}{2005}\natexlab{}.
\newblock \bibinfo{booktitle}{\emph{Robust regression and outlier detection}}.
\newblock \bibinfo{publisher}{John wiley \& sons}.
\newblock


\bibitem[Ruff et~al\mbox{.}(2021)]%
        {ruff2021unifying}
\bibfield{author}{\bibinfo{person}{Lukas Ruff}, \bibinfo{person}{Jacob~R
  Kauffmann}, \bibinfo{person}{Robert~A Vandermeulen},
  \bibinfo{person}{Gr{\'e}goire Montavon}, \bibinfo{person}{Wojciech Samek},
  \bibinfo{person}{Marius Kloft}, \bibinfo{person}{Thomas~G Dietterich}, {and}
  \bibinfo{person}{Klaus-Robert M{\"u}ller}.} \bibinfo{year}{2021}\natexlab{}.
\newblock \showarticletitle{A unifying review of deep and shallow anomaly
  detection}.
\newblock \bibinfo{journal}{\emph{Proc. IEEE}} \bibinfo{volume}{109},
  \bibinfo{number}{5} (\bibinfo{year}{2021}), \bibinfo{pages}{756--795}.
\newblock


\bibitem[Ruff et~al\mbox{.}(2018)]%
        {ruff2018deep}
\bibfield{author}{\bibinfo{person}{Lukas Ruff}, \bibinfo{person}{Robert
  Vandermeulen}, \bibinfo{person}{Nico Goernitz}, \bibinfo{person}{Lucas
  Deecke}, \bibinfo{person}{Shoaib~Ahmed Siddiqui}, \bibinfo{person}{Alexander
  Binder}, \bibinfo{person}{Emmanuel M{\"u}ller}, {and} \bibinfo{person}{Marius
  Kloft}.} \bibinfo{year}{2018}\natexlab{}.
\newblock \showarticletitle{Deep one-class classification}. In
  \bibinfo{booktitle}{\emph{International conference on machine learning}}.
  PMLR, \bibinfo{pages}{4393--4402}.
\newblock


\bibitem[Sakurada and Yairi(2014)]%
        {sakurada2014anomaly}
\bibfield{author}{\bibinfo{person}{Mayu Sakurada} {and}
  \bibinfo{person}{Takehisa Yairi}.} \bibinfo{year}{2014}\natexlab{}.
\newblock \showarticletitle{Anomaly detection using autoencoders with nonlinear
  dimensionality reduction}. In \bibinfo{booktitle}{\emph{Proceedings of the
  MLSDA 2014 2nd workshop on machine learning for sensory data analysis}}.
  \bibinfo{pages}{4--11}.
\newblock


\bibitem[Salinas et~al\mbox{.}(2020)]%
        {salinas2020deepar}
\bibfield{author}{\bibinfo{person}{David Salinas}, \bibinfo{person}{Valentin
  Flunkert}, \bibinfo{person}{Jan Gasthaus}, {and} \bibinfo{person}{Tim
  Januschowski}.} \bibinfo{year}{2020}\natexlab{}.
\newblock \showarticletitle{DeepAR: Probabilistic forecasting with
  autoregressive recurrent networks}.
\newblock \bibinfo{journal}{\emph{International Journal of Forecasting}}
  \bibinfo{volume}{36}, \bibinfo{number}{3} (\bibinfo{year}{2020}),
  \bibinfo{pages}{1181--1191}.
\newblock


\bibitem[Schmidl et~al\mbox{.}(2022)]%
        {schmidl2022anomaly}
\bibfield{author}{\bibinfo{person}{Sebastian Schmidl}, \bibinfo{person}{Phillip
  Wenig}, {and} \bibinfo{person}{Thorsten Papenbrock}.}
  \bibinfo{year}{2022}\natexlab{}.
\newblock \showarticletitle{Anomaly detection in time series: a comprehensive
  evaluation}.
\newblock \bibinfo{journal}{\emph{Proceedings of the VLDB Endowment}}
  \bibinfo{volume}{15}, \bibinfo{number}{9} (\bibinfo{year}{2022}),
  \bibinfo{pages}{1779--1797}.
\newblock


\bibitem[Shen et~al\mbox{.}(2020)]%
        {shen2020timeseries}
\bibfield{author}{\bibinfo{person}{Lifeng Shen}, \bibinfo{person}{Zhuocong Li},
  {and} \bibinfo{person}{James Kwok}.} \bibinfo{year}{2020}\natexlab{}.
\newblock \showarticletitle{Timeseries anomaly detection using temporal
  hierarchical one-class network}.
\newblock \bibinfo{journal}{\emph{Advances in Neural Information Processing
  Systems}}  \bibinfo{volume}{33} (\bibinfo{year}{2020}),
  \bibinfo{pages}{13016--13026}.
\newblock


\bibitem[Shin et~al\mbox{.}(2020)]%
        {shin2020itad}
\bibfield{author}{\bibinfo{person}{Youjin Shin}, \bibinfo{person}{Sangyup Lee},
  \bibinfo{person}{Shahroz Tariq}, \bibinfo{person}{Myeong~Shin Lee},
  \bibinfo{person}{Okchul Jung}, \bibinfo{person}{Daewon Chung}, {and}
  \bibinfo{person}{Simon~S Woo}.} \bibinfo{year}{2020}\natexlab{}.
\newblock \showarticletitle{Itad: integrative tensor-based anomaly detection
  system for reducing false positives of satellite systems}. In
  \bibinfo{booktitle}{\emph{Proceedings of the 29th ACM international
  conference on information \& knowledge management}}.
  \bibinfo{pages}{2733--2740}.
\newblock


\bibitem[Su et~al\mbox{.}(2019)]%
        {su2019robust}
\bibfield{author}{\bibinfo{person}{Ya Su}, \bibinfo{person}{Youjian Zhao},
  \bibinfo{person}{Chenhao Niu}, \bibinfo{person}{Rong Liu},
  \bibinfo{person}{Wei Sun}, {and} \bibinfo{person}{Dan Pei}.}
  \bibinfo{year}{2019}\natexlab{}.
\newblock \showarticletitle{Robust anomaly detection for multivariate time
  series through stochastic recurrent neural network}. In
  \bibinfo{booktitle}{\emph{Proceedings of the 25th ACM SIGKDD international
  conference on knowledge discovery \& data mining}}.
  \bibinfo{pages}{2828--2837}.
\newblock


\bibitem[Tariq et~al\mbox{.}(2019)]%
        {tariq2019detecting}
\bibfield{author}{\bibinfo{person}{Shahroz Tariq}, \bibinfo{person}{Sangyup
  Lee}, \bibinfo{person}{Youjin Shin}, \bibinfo{person}{Myeong~Shin Lee},
  \bibinfo{person}{Okchul Jung}, \bibinfo{person}{Daewon Chung}, {and}
  \bibinfo{person}{Simon~S Woo}.} \bibinfo{year}{2019}\natexlab{}.
\newblock \showarticletitle{Detecting anomalies in space using multivariate
  convolutional LSTM with mixtures of probabilistic PCA}. In
  \bibinfo{booktitle}{\emph{Proceedings of the 25th ACM SIGKDD international
  conference on knowledge discovery \& data mining}}.
  \bibinfo{pages}{2123--2133}.
\newblock


\bibitem[Tax and Duin(2004)]%
        {tax2004support}
\bibfield{author}{\bibinfo{person}{David~MJ Tax} {and}
  \bibinfo{person}{Robert~PW Duin}.} \bibinfo{year}{2004}\natexlab{}.
\newblock \showarticletitle{Support vector data description}.
\newblock \bibinfo{journal}{\emph{Machine learning}} \bibinfo{volume}{54},
  \bibinfo{number}{1} (\bibinfo{year}{2004}), \bibinfo{pages}{45--66}.
\newblock


\bibitem[Ulyanov et~al\mbox{.}(2017)]%
        {ulyanov2017improved}
\bibfield{author}{\bibinfo{person}{Dmitry Ulyanov}, \bibinfo{person}{Andrea
  Vedaldi}, {and} \bibinfo{person}{Victor Lempitsky}.}
  \bibinfo{year}{2017}\natexlab{}.
\newblock \showarticletitle{Improved texture networks: Maximizing quality and
  diversity in feed-forward stylization and texture synthesis}. In
  \bibinfo{booktitle}{\emph{Proceedings of the IEEE conference on computer
  vision and pattern recognition}}. \bibinfo{pages}{6924--6932}.
\newblock


\bibitem[Vaswani et~al\mbox{.}(2017)]%
        {vaswani2017attention}
\bibfield{author}{\bibinfo{person}{Ashish Vaswani}, \bibinfo{person}{Noam
  Shazeer}, \bibinfo{person}{Niki Parmar}, \bibinfo{person}{Jakob Uszkoreit},
  \bibinfo{person}{Llion Jones}, \bibinfo{person}{Aidan~N Gomez},
  \bibinfo{person}{{\L}ukasz Kaiser}, {and} \bibinfo{person}{Illia
  Polosukhin}.} \bibinfo{year}{2017}\natexlab{}.
\newblock \showarticletitle{Attention is all you need}.
\newblock \bibinfo{journal}{\emph{Advances in neural information processing
  systems}}  \bibinfo{volume}{30} (\bibinfo{year}{2017}).
\newblock


\bibitem[Wen et~al\mbox{.}(2022)]%
        {wen2022robust}
\bibfield{author}{\bibinfo{person}{Qingsong Wen}, \bibinfo{person}{Linxiao
  Yang}, \bibinfo{person}{Tian Zhou}, {and} \bibinfo{person}{Liang Sun}.}
  \bibinfo{year}{2022}\natexlab{}.
\newblock \showarticletitle{Robust time series analysis and applications: An
  industrial perspective}. In \bibinfo{booktitle}{\emph{Proceedings of the 28th
  ACM SIGKDD Conference on Knowledge Discovery and Data Mining}}.
  \bibinfo{pages}{4836--4837}.
\newblock


\bibitem[Wu et~al\mbox{.}(2021)]%
        {wu2021autoformer}
\bibfield{author}{\bibinfo{person}{Haixu Wu}, \bibinfo{person}{Jiehui Xu},
  \bibinfo{person}{Jianmin Wang}, {and} \bibinfo{person}{Mingsheng Long}.}
  \bibinfo{year}{2021}\natexlab{}.
\newblock \showarticletitle{Autoformer: Decomposition transformers with
  auto-correlation for long-term series forecasting}.
\newblock \bibinfo{journal}{\emph{Advances in Neural Information Processing
  Systems}}  \bibinfo{volume}{34} (\bibinfo{year}{2021}),
  \bibinfo{pages}{22419--22430}.
\newblock


\bibitem[Wu et~al\mbox{.}(2018)]%
        {wu2018unsupervised}
\bibfield{author}{\bibinfo{person}{Zhirong Wu}, \bibinfo{person}{Yuanjun
  Xiong}, \bibinfo{person}{Stella~X Yu}, {and} \bibinfo{person}{Dahua Lin}.}
  \bibinfo{year}{2018}\natexlab{}.
\newblock \showarticletitle{Unsupervised feature learning via non-parametric
  instance discrimination}. In \bibinfo{booktitle}{\emph{Proceedings of the
  IEEE conference on computer vision and pattern recognition}}.
  \bibinfo{pages}{3733--3742}.
\newblock


\bibitem[Xu et~al\mbox{.}(2018)]%
        {xu2018unsupervised}
\bibfield{author}{\bibinfo{person}{Haowen Xu}, \bibinfo{person}{Wenxiao Chen},
  \bibinfo{person}{Nengwen Zhao}, \bibinfo{person}{Zeyan Li},
  \bibinfo{person}{Jiahao Bu}, \bibinfo{person}{Zhihan Li},
  \bibinfo{person}{Ying Liu}, \bibinfo{person}{Youjian Zhao},
  \bibinfo{person}{Dan Pei}, \bibinfo{person}{Yang Feng}, {et~al\mbox{.}}}
  \bibinfo{year}{2018}\natexlab{}.
\newblock \showarticletitle{Unsupervised anomaly detection via variational
  auto-encoder for seasonal kpis in web applications}. In
  \bibinfo{booktitle}{\emph{Proceedings of the 2018 world wide web
  conference}}. \bibinfo{pages}{187--196}.
\newblock


\bibitem[Xu et~al\mbox{.}(2021)]%
        {xu2021anomaly}
\bibfield{author}{\bibinfo{person}{Jiehui Xu}, \bibinfo{person}{Haixu Wu},
  \bibinfo{person}{Jianmin Wang}, {and} \bibinfo{person}{Mingsheng Long}.}
  \bibinfo{year}{2021}\natexlab{}.
\newblock \showarticletitle{Anomaly transformer: Time series anomaly detection
  with association discrepancy}.
\newblock \bibinfo{journal}{\emph{arXiv preprint arXiv:2110.02642}}
  (\bibinfo{year}{2021}).
\newblock


\bibitem[Yairi et~al\mbox{.}(2017)]%
        {yairi2017data}
\bibfield{author}{\bibinfo{person}{Takehisa Yairi}, \bibinfo{person}{Naoya
  Takeishi}, \bibinfo{person}{Tetsuo Oda}, \bibinfo{person}{Yuta Nakajima},
  \bibinfo{person}{Naoki Nishimura}, {and} \bibinfo{person}{Noboru Takata}.}
  \bibinfo{year}{2017}\natexlab{}.
\newblock \showarticletitle{A data-driven health monitoring method for
  satellite housekeeping data based on probabilistic clustering and
  dimensionality reduction}.
\newblock \bibinfo{journal}{\emph{IEEE Trans. Aerospace Electron. Systems}}
  \bibinfo{volume}{53}, \bibinfo{number}{3} (\bibinfo{year}{2017}),
  \bibinfo{pages}{1384--1401}.
\newblock


\bibitem[Yang et~al\mbox{.}(2023)]%
        {yang2023sgdp}
\bibfield{author}{\bibinfo{person}{Yiyuan Yang}, \bibinfo{person}{Rongshang
  Li}, \bibinfo{person}{Qiquan Shi}, \bibinfo{person}{Xijun Li},
  \bibinfo{person}{Gang Hu}, \bibinfo{person}{Xing Li}, {and}
  \bibinfo{person}{Mingxuan Yuan}.} \bibinfo{year}{2023}\natexlab{}.
\newblock \showarticletitle{SGDP: A Stream-Graph Neural Network Based Data
  Prefetcher}.
\newblock \bibinfo{journal}{\emph{arXiv preprint arXiv:2304.03864}}
  (\bibinfo{year}{2023}).
\newblock


\bibitem[Yang et~al\mbox{.}(2021a)]%
        {yang2021pipeline2}
\bibfield{author}{\bibinfo{person}{Yiyuan Yang}, \bibinfo{person}{Yi Li}, {and}
  \bibinfo{person}{Haifeng Zhang}.} \bibinfo{year}{2021}\natexlab{a}.
\newblock \showarticletitle{Pipeline safety early warning method for
  distributed signal using bilinear CNN and LightGBM}. In
  \bibinfo{booktitle}{\emph{ICASSP 2021-2021 IEEE International Conference on
  Acoustics, Speech and Signal Processing (ICASSP)}}. IEEE,
  \bibinfo{pages}{4110--4114}.
\newblock


\bibitem[Yang et~al\mbox{.}(2021b)]%
        {yang2021early}
\bibfield{author}{\bibinfo{person}{Yiyuan Yang}, \bibinfo{person}{Yi Li},
  \bibinfo{person}{Taojia Zhang}, \bibinfo{person}{Yan Zhou}, {and}
  \bibinfo{person}{Haifeng Zhang}.} \bibinfo{year}{2021}\natexlab{b}.
\newblock \showarticletitle{Early safety warnings for long-distance pipelines:
  A distributed optical fiber sensor machine learning approach}. In
  \bibinfo{booktitle}{\emph{Proceedings of the AAAI Conference on Artificial
  Intelligence}}, Vol.~\bibinfo{volume}{35}. \bibinfo{pages}{14991--14999}.
\newblock


\bibitem[Yang et~al\mbox{.}(2021c)]%
        {yang2021long}
\bibfield{author}{\bibinfo{person}{Yiyuan Yang}, \bibinfo{person}{Haifeng
  Zhang}, {and} \bibinfo{person}{Yi Li}.} \bibinfo{year}{2021}\natexlab{c}.
\newblock \showarticletitle{Long-distance pipeline safety early warning: a
  distributed optical fiber sensing semi-supervised learning method}.
\newblock \bibinfo{journal}{\emph{IEEE Sensors Journal}} \bibinfo{volume}{21},
  \bibinfo{number}{17} (\bibinfo{year}{2021}), \bibinfo{pages}{19453--19461}.
\newblock


\bibitem[Yang et~al\mbox{.}(2021d)]%
        {yang2021pipeline}
\bibfield{author}{\bibinfo{person}{Yiyuan Yang}, \bibinfo{person}{Haifeng
  Zhang}, {and} \bibinfo{person}{Yi Li}.} \bibinfo{year}{2021}\natexlab{d}.
\newblock \showarticletitle{Pipeline safety early warning by
  multifeature-fusion CNN and LightGBM analysis of signals from distributed
  optical fiber sensors}.
\newblock \bibinfo{journal}{\emph{IEEE Transactions on Instrumentation and
  Measurement}}  \bibinfo{volume}{70} (\bibinfo{year}{2021}),
  \bibinfo{pages}{1--13}.
\newblock


\bibitem[Ye et~al\mbox{.}(2019)]%
        {ye2019unsupervised}
\bibfield{author}{\bibinfo{person}{Mang Ye}, \bibinfo{person}{Xu Zhang},
  \bibinfo{person}{Pong~C Yuen}, {and} \bibinfo{person}{Shih-Fu Chang}.}
  \bibinfo{year}{2019}\natexlab{}.
\newblock \showarticletitle{Unsupervised embedding learning via invariant and
  spreading instance feature}. In \bibinfo{booktitle}{\emph{Proceedings of the
  IEEE/CVF Conference on Computer Vision and Pattern Recognition}}.
  \bibinfo{pages}{6210--6219}.
\newblock


\bibitem[Yeh et~al\mbox{.}(2016)]%
        {yeh2016matrix}
\bibfield{author}{\bibinfo{person}{Chin-Chia~Michael Yeh}, \bibinfo{person}{Yan
  Zhu}, \bibinfo{person}{Liudmila Ulanova}, \bibinfo{person}{Nurjahan Begum},
  \bibinfo{person}{Yifei Ding}, \bibinfo{person}{Hoang~Anh Dau},
  \bibinfo{person}{Diego~Furtado Silva}, \bibinfo{person}{Abdullah Mueen},
  {and} \bibinfo{person}{Eamonn Keogh}.} \bibinfo{year}{2016}\natexlab{}.
\newblock \showarticletitle{Matrix profile I: all pairs similarity joins for
  time series: a unifying view that includes motifs, discords and shapelets}.
  In \bibinfo{booktitle}{\emph{2016 IEEE 16th international conference on data
  mining (ICDM)}}. Ieee, \bibinfo{pages}{1317--1322}.
\newblock


\bibitem[Yu and Sun(2020)]%
        {yu2020policy}
\bibfield{author}{\bibinfo{person}{Mengran Yu} {and} \bibinfo{person}{Shiliang
  Sun}.} \bibinfo{year}{2020}\natexlab{}.
\newblock \showarticletitle{Policy-based reinforcement learning for time series
  anomaly detection}.
\newblock \bibinfo{journal}{\emph{Engineering Applications of Artificial
  Intelligence}}  \bibinfo{volume}{95} (\bibinfo{year}{2020}),
  \bibinfo{pages}{103919}.
\newblock


\bibitem[Zamanzadeh~Darban et~al\mbox{.}(2022)]%
        {zamanzadeh2022deep}
\bibfield{author}{\bibinfo{person}{Zahra Zamanzadeh~Darban},
  \bibinfo{person}{Geoffrey~I Webb}, \bibinfo{person}{Shirui Pan},
  \bibinfo{person}{Charu~C Aggarwal}, {and} \bibinfo{person}{Mahsa Salehi}.}
  \bibinfo{year}{2022}\natexlab{}.
\newblock \showarticletitle{Deep Learning for Time Series Anomaly Detection: A
  Survey}.
\newblock \bibinfo{journal}{\emph{arXiv e-prints}} (\bibinfo{year}{2022}),
  \bibinfo{pages}{arXiv--2211}.
\newblock


\bibitem[Zhang et~al\mbox{.}(2022b)]%
        {zhang2022does}
\bibfield{author}{\bibinfo{person}{Chaoning Zhang}, \bibinfo{person}{Kang
  Zhang}, \bibinfo{person}{Chenshuang Zhang}, \bibinfo{person}{Trung~X Pham},
  \bibinfo{person}{Chang~D Yoo}, {and} \bibinfo{person}{In~So Kweon}.}
  \bibinfo{year}{2022}\natexlab{b}.
\newblock \showarticletitle{How does {SimSiam} avoid collapse without negative
  samples? a unified understanding with self-supervised contrastive learning}.
\newblock \bibinfo{journal}{\emph{Proceedings of International Conference on
  Learning Representations (ICLR)}} (\bibinfo{year}{2022}).
\newblock


\bibitem[Zhang et~al\mbox{.}(2022c)]%
        {zhang2022tfad}
\bibfield{author}{\bibinfo{person}{Chaoli Zhang}, \bibinfo{person}{Tian Zhou},
  \bibinfo{person}{Qingsong Wen}, {and} \bibinfo{person}{Liang Sun}.}
  \bibinfo{year}{2022}\natexlab{c}.
\newblock \showarticletitle{TFAD: A Decomposition Time Series Anomaly Detection
  Architecture with Time-Frequency Analysis}. In
  \bibinfo{booktitle}{\emph{Proceedings of the 31st ACM International
  Conference on Information \& Knowledge Management}}.
  \bibinfo{pages}{2497--2507}.
\newblock


\bibitem[Zhang et~al\mbox{.}(2022a)]%
        {zhang2022adaptive}
\bibfield{author}{\bibinfo{person}{Yuxin Zhang}, \bibinfo{person}{Jindong
  Wang}, \bibinfo{person}{Yiqiang Chen}, \bibinfo{person}{Han Yu}, {and}
  \bibinfo{person}{Tao Qin}.} \bibinfo{year}{2022}\natexlab{a}.
\newblock \showarticletitle{Adaptive memory networks with self-supervised
  learning for unsupervised anomaly detection}.
\newblock \bibinfo{journal}{\emph{IEEE Transactions on Knowledge and Data
  Engineering}} (\bibinfo{year}{2022}).
\newblock


\bibitem[Zhao et~al\mbox{.}(2020a)]%
        {Zhao2020MultivariateTA}
\bibfield{author}{\bibinfo{person}{Hang Zhao}, \bibinfo{person}{Yujing Wang},
  \bibinfo{person}{Juanyong Duan}, \bibinfo{person}{Congrui Huang},
  \bibinfo{person}{Defu Cao}, \bibinfo{person}{Yunhai Tong},
  \bibinfo{person}{Bixiong Xu}, \bibinfo{person}{Jing Bai},
  \bibinfo{person}{Jie Tong}, {and} \bibinfo{person}{Qi Zhang}.}
  \bibinfo{year}{2020}\natexlab{a}.
\newblock \showarticletitle{Multivariate Time-series Anomaly Detection via
  Graph Attention Network}.
\newblock \bibinfo{journal}{\emph{2020 IEEE International Conference on Data
  Mining (ICDM)}} (\bibinfo{year}{2020}), \bibinfo{pages}{841--850}.
\newblock


\bibitem[Zhao et~al\mbox{.}(2020b)]%
        {zhao2020multivariate}
\bibfield{author}{\bibinfo{person}{Hang Zhao}, \bibinfo{person}{Yujing Wang},
  \bibinfo{person}{Juanyong Duan}, \bibinfo{person}{Congrui Huang},
  \bibinfo{person}{Defu Cao}, \bibinfo{person}{Yunhai Tong},
  \bibinfo{person}{Bixiong Xu}, \bibinfo{person}{Jing Bai},
  \bibinfo{person}{Jie Tong}, {and} \bibinfo{person}{Qi Zhang}.}
  \bibinfo{year}{2020}\natexlab{b}.
\newblock \showarticletitle{Multivariate time-series anomaly detection via
  graph attention network}. In \bibinfo{booktitle}{\emph{2020 IEEE
  International Conference on Data Mining (ICDM)}}. IEEE,
  \bibinfo{pages}{841--850}.
\newblock


\bibitem[Zhou et~al\mbox{.}(2019)]%
        {zhou2019beatgan}
\bibfield{author}{\bibinfo{person}{Bin Zhou}, \bibinfo{person}{Shenghua Liu},
  \bibinfo{person}{Bryan Hooi}, \bibinfo{person}{Xueqi Cheng}, {and}
  \bibinfo{person}{Jing Ye}.} \bibinfo{year}{2019}\natexlab{}.
\newblock \showarticletitle{BeatGAN: Anomalous Rhythm Detection using
  Adversarially Generated Time Series.}. In \bibinfo{booktitle}{\emph{IJCAI}},
  Vol.~\bibinfo{volume}{2019}. \bibinfo{pages}{4433--4439}.
\newblock


\bibitem[Zong et~al\mbox{.}(2018)]%
        {zong2018deep}
\bibfield{author}{\bibinfo{person}{Bo Zong}, \bibinfo{person}{Qi Song},
  \bibinfo{person}{Martin~Renqiang Min}, \bibinfo{person}{Wei Cheng},
  \bibinfo{person}{Cristian Lumezanu}, \bibinfo{person}{Daeki Cho}, {and}
  \bibinfo{person}{Haifeng Chen}.} \bibinfo{year}{2018}\natexlab{}.
\newblock \showarticletitle{Deep autoencoding gaussian mixture model for
  unsupervised anomaly detection}. In \bibinfo{booktitle}{\emph{International
  conference on learning representations}}.
\newblock


\end{thebibliography}

\newpage

\appendix

\newpage
\section{Algorithms}\label{sec:appendix}
\begin{algorithm}[htb]
	\caption{DCdetector Outliner}
	\label{alg:code1}

	\definecolor{codeblue}{rgb}{0.25,0.5,0.5}
	\lstset{
		backgroundcolor=\color{white},
		basicstyle=\fontsize{7.2pt}{7.2pt}\ttfamily\selectfont,
		columns=fullflexible,
		breaklines=true,
		captionpos=b,
		commentstyle=\fontsize{7.2pt}{7.2pt}\color{codeblue},
		keywordstyle=\fontsize{7.2pt}{7.2pt},
	}
    
    \begin{lstlisting}[language=python]
from einops import rearrange

class DCdetector(nn.Module):
    def __init__(self, win_size, enc_in, c_out, n_heads, d_model, e_layers, patch_size, channel, d_ff, dropout):
        super(DCdetector, self).__init__()
        self.patch_size = patch_size
        self.channel = channel
        self.win_size = win_size
        
        # Patching List Embedding  
        self.embedding_patch_size = nn.ModuleList()
        self.embedding_patch_num = nn.ModuleList()
        for i, patchsize in enumerate(self.patch_size):
            self.embedding_patch_size.append(DataEmbedding(patchsize, d_model, dropout))
            self.embedding_patch_num.append(DataEmbedding(self.win_size//patchsize, d_model, dropout))
        self.embedding_window_size = DataEmbedding(enc_in, d_model, dropout)
         
        # Dual Attention Encoder
        self.encoder = Encoder(
            [AttentionLayer(
                    DAC_structure(win_size, patch_size, channel, False, attention_dropout=dropout, output_attention=output_attention),
                d_model, patch_size, channel, n_heads, win_size) 
                for l in range(e_layers)
            ],norm_layer=torch.nn.LayerNorm(d_model))

    def forward(self, x):
        B, L, M = x.shape #Batch win_size channel
        revin_layer = RevIN(num_features=M)
        patch_wise_mean = [], in_patch_mean = []
        x_ori = self.embedding_window_size(x)
          
        # Instance Normalization Operation
        x = revin_layer(x, 'norm')     
        
        # Mutil-scale Patching Operation 
        for patch_index, patchsize in enumerate(self.patch_size):   
            x_patch_size = rearrange(x, 'b (n p) m -> (b m) n p', p = patchsize) 
            x_patch_num = rearrange(x, 'b (p n) m -> (b m) p n', p = patchsize) 
            x_patch_size = self.embedding_patch_size[patch_index](x_patch_size)
            x_patch_num = self.embedding_patch_num[patch_index](x_patch_num)
            patch_wise, in_patch = self.encoder(x_patch_size, x_patch_num, x_ori, patch_index)
            patch_wise_mean.append(patch_wise)
            in_patch_mean.append(in_patch)
            
        return patch_wise_mean, in_patch_mean
        
    \end{lstlisting}
\end{algorithm}

\begin{algorithm}[htb]
	\caption{Dual Attention Contrastive Structure}
	\label{alg:code2}
	\definecolor{codeblue}{rgb}{0.25,0.5,0.5}
	\lstset{
		backgroundcolor=\color{white},
		basicstyle=\fontsize{7.2pt}{7.2pt}\ttfamily\selectfont,
		columns=fullflexible,
		breaklines=true,
		captionpos=b,
		commentstyle=\fontsize{7.2pt}{7.2pt}\color{codeblue},
		keywordstyle=\fontsize{7.2pt}{7.2pt},
	}
    
    \begin{lstlisting}[language=python]
from einops import reduce, repeat
from math import sqrt

class DAC_structure(nn.Module):
    def __init__(self, win_size, patch_size, channel, scale=None, attention_dropout=0.05):
        super(DAC_structure, self).__init__()
        self.scale = scale
        self.dropout = nn.Dropout(attention_dropout)
        self.window_size = win_size
        self.patch_size = patch_size
        self.channel = channel

    def forward(self, queries_patch_size, queries_patch_num, keys_patch_size, keys_patch_num, values, patch_index, attn_mask):
                                                 
        # Patch-wise Representation
        B, L, H, E = queries_patch_size.shape #batch_size*channel, patch_num, n_head, d_model/n_head
        scale_patch_size = self.scale or 1. / sqrt(E)
        scores_patch_size = torch.einsum("blhe,bshe->bhls", queries_patch_size, keys_patch_size) #batch*ch, nheads, p_num, p_num   
        attn_patch_size = scale_patch_size * scores_patch_size
        series_patch_size = self.dropout(torch.softmax(attn_patch_size, dim=-1)) # B*D_model H N N

        # In-patch Representation
        B, L, H, E = queries_patch_num.shape #batch_size*channel, patch_size, n_head, d_model/n_head
        scale_patch_num = self.scale or 1. / sqrt(E)
        scores_patch_num = torch.einsum("blhe,bshe->bhls", queries_patch_num, keys_patch_num) #batch*ch, nheads, p_size, p_size 
        attn_patch_num = scale_patch_num * scores_patch_num
        series_patch_num = self.dropout(torch.softmax(attn_patch_num, dim=-1)) # B*D_model H S S 

        # Upsampling
        series_patch_size = repeat(series_patch_size, 'blmn->bl(m repeat_m)(n repeat_n)', repeat_m=self.patch_size[patch_index], repeat_n=self.patch_size[patch_index])  
        series_patch_num = series_patch_num.repeat(1,1,self.window_size//self.patch_size[patch_index],self.window_size//self.patch_size[patch_index]) 

        return series_patch_size, series_patch_num
    \end{lstlisting}
\end{algorithm}

Two main algorithms and their codes are presented as follows, the DCdetector Outliner (Algorithm~\ref{alg:code1}) and the Dual Attention Contrastive Structure (Algorithm~\ref{alg:code2}). 

For the DCdetector Outliner (Algorithm~\ref{alg:code1}), we use nn.ModuleList() in PyTorch~\cite{paszke2019pytorch} to define multiple patching scales and perform embedding and dual attention operations for each set of patch sizes. After instance normalization, patch-wise encoders and in-patch encoders are calculated in the patch number $N$ dimension and the patch size $P$ dimension, respectively. Finally, we average the different patching scales to obtain the final patch-wise representation $\mathcal{N}$ and in-patch representation $\mathcal{P}$.

For the Dual Attention Contrastive Structure (Algorithm~\ref{alg:code2}), we calculate patch-wise representation based on Eq.\ref{eq1} - Eq.\ref{eq3} and in-patch representation using Eq.\ref{eq4} - Eq.\ref{eq6}, respectively. Finally, we apply different upsampling methods to make the shape of the two representations the same for subsequent comparison of representation discrepancy. Note that, we do not need to calculate the specific attention values, as only two representations are made and the attention weights can also be used as representations as well as improving the efficiency of the code. Besides, only a single patch scale of dual attention contrastive structure is shown here, which may suffer from information loss when upsampling is performed. However, the multi-patch scale will compensate for this issue, as shown in Algorithm~\ref{alg:code1}.

\section{Dataset Description}\label{sec:datasets}

We summarize the seven adopted benchmark datasets for evaluation in Table \ref{Tab: Dataset Description}. These datasets include both univariate and multivariate time series scenarios with different types and anomaly ratios. MSL, SMAP, PSM, SMD, SWaT, NIPS-TS-SWAN, and NIPS-TS-GECCO are multivariate time series datasets. UCR is a univariate time series dataset.

\section{Extra Studies}\label{sec:appendix_extra_ablation}
To verify the sensitivity of the parameters in the proposed DCdetector, more ablation experiments are conducted in this part. We provide more detailed results here than those in Section~\ref{sec:ablation}. We also show the memory used as well as the iteration time spent during the training process.

\subsection{Study on Metrics in Loss Function}
We use different statistical distances to calculate the discrepancy between patch-wise representation and in-patch representation, and the results are shown in Table~\ref{Tab: Ablation Metric Loss Function}. The loss function proposed in Section~\ref{sec:rd} can get the SOTA performance in all benchmarks. Note that only using simple KL divergence, which is an asymmetrical loss function, we can still get a comparable result. However, for the Jensen-Shannon (JS) divergence, there is visible performance degradation, especially for the MSL benchmark.

\subsection{Study on Multi-scale Patching}
The multi-patching scale $\in\{[1],[3],[5],[1,3],[1,5],[3,5],[1,3,5]\}$ are tested. Patch size preference is for odd numbers to prevent information loss during upsampling. Generally, multi-scale design results in larger memory and different datasets have different best multi-patching scales. This is perhaps due to different information densities and anomaly types in different situations. The details of evaluation results are shown in Table~\ref{Tab: Ablation multiscale results}.

\subsection{Study on Window Size}
Window size is a significant hyper-parameter in time series analysis. It is used to split time series into instances, as usually, a single point can not be considered as a sample. The results in Table~\ref{Tab: Ablation window size results} show that DCdetector is rather robust in different window sizes. Actually, in a large range [45, 195], the performances are slightly lower than the best ones for all benchmarks. Besides, we also test the impact of window size on memory cost and running time. The window size will affect the memory cost in a quadratic computational complexity way. So, the trade-off between slide window size and memory cost/running time is pretty important, especially for real-life scenarios. Fortunately, DCdetector can work optimally with a window size of less than 105 in all benchmarks, which greatly decreases the complexity of the model and its memory cost.

\subsection{Study on Attention Head}
Generally, multi-head attention is widely used in attention networks. We study the influence of attention head number $H$ in DCdetector. In general, the number of attention heads is even, so we set $H \in \{1,2,4,8\}$. Fortunately, with a small attention head number, as shown in Table~\ref{Tab: Ablation attention head number results}, our model still achieves good performances (the best one or slightly lower than the best). Thus, DCdetector does not need a large memory when running. 

\subsection{Study on Embedding Dimension}
The embedding dimension $d_{model}$ is another important parameter in the attention network. As a hyperparameter of the hidden channels, it may have impacts on model performance, memory cost, and running efficiency. We set $d_{model} \in \{128,256,512,1024\}$ as suggested hyperparameters by Transformer \cite{vaswani2017attention}. For SMAP and PSM, it has little effect on the final results. As for MSL, it achieves the best performance with a small $d_{model}$ size and small memory. Overall, the proposed DCdetector can achieve quite good performance even with a small memory cost and good real-time performance. Details are in Table~\ref{Tab: Ablation Embedding d_model results}.

\subsection{Study on Encoder Layer}
Many deep models' performances are dependent on the number of network layers $L$. We also show the influence of the number of encoder layers in Table~\ref{Tab: Ablation Encoder layers results}. We set $L \in \{1,2,3,4,5\}$ as suggested hyperparameters by Transformer \cite{vaswani2017attention}. Different benchmarks have different optimal parameters. Luckily, our model can gain the best performance in no more than 3 layers, and will not fail with too few encoder layers or over-fit with too many encoder layers.

\subsection{Study on Anomaly Threshold}
Anomaly threshold $\delta$ is a hyperparameter, which may affect the determination of anomaly or not, based on Eq. \ref{eq11}. We have a default value of 1 for all benchmarks. As shown in Table~\ref{Tab: Ablation Anomaly Threshold results}, when it is in the range of 0.5 to 1, it has little effect on the final model performance. PSM and SMAP are also more robust to anomaly threshold than MSL. For the three benchmarks, its best results appear when $\delta$ equals 0.7 or 0.8.

\begin{table*}[!t]
\caption{Details of benchmark datasets. AR (anomaly ratio) represents the abnormal proportion of the whole dataset.}
{
\begin{tabular}{c|ccccccc}
\hline \hline
\textbf{Benchmark}     & \textbf{Source}                  & \textbf{Dimension} & \textbf{Window} & \textbf{Patch Size}  & \textbf{\#Training} & \textbf{\#Test (Labeled)} & \textbf{AR (\%)} \\ \hline
MSL           & NASA Space Sensors      & 55        & 90     & {[}3,5{]}   & 58,317     & 73,729           & 10.5    \\
SMAP          & NASA Space Sensors      & 25        & 105    & {[}3,5,7{]} & 135,183    & 427,617          & 12.8    \\
PSM           & eBay Server Machine     & 25        & 60     & {[}1,3,5{]} & 132,481    & 87,841           & 27.8    \\
SMD           & Internet Server Machine & 38        & 105    & {[}5,7{]}   & 708,405    & 708,420          & 4.2     \\
SWaT           & Infrastructure System & 51        & 105    & {[}3,5,7{]}   & 495,000    &  449,919         & 12.1     \\
NIPS-TS-SWAN  & Space (Solar) Weather   & 38        & 36     & {[}1,3{]}   & 60,000     & 60,000           & 32.6    \\
NIPS-TS-GECCO & Water Quality for IoT   & 9         & 90     & {[}1,3,5{]} & 69,260     & 69,261           & 1.1     \\ \hline
UCR           & Various Natural Sources & 1         & 105    & {[}3,5,7{]} & 2,238,349  & 6,143,541        & 0.6     \\\hline \hline
\end{tabular}}
\label{Tab: Dataset Description}
\end{table*}

\begin{table*}[!ht]
\caption{Ablation studies on metrics in the loss function. All results are in \%. The best ones are in \textbf{Bold}.}
{
\begin{tabular}{c|lll|lll|lll}
\hline \hline
\textbf{Dataset} & \multicolumn{3}{c|}{\textbf{MSL}}                                                                  & \multicolumn{3}{c|}{\textbf{SMAP}}                                                                 & \multicolumn{3}{c}{\textbf{PSM}}                                                                  \\ \hline
\textbf{Metric}  & \multicolumn{1}{c}{\textbf{P}} & \multicolumn{1}{c}{\textbf{R}} & \multicolumn{1}{c|}{\textbf{F1}} & \multicolumn{1}{c}{\textbf{P}} & \multicolumn{1}{c}{\textbf{R}} & \multicolumn{1}{c|}{\textbf{F1}} & \multicolumn{1}{c}{\textbf{P}} & \multicolumn{1}{c}{\textbf{R}} & \multicolumn{1}{c}{\textbf{F1}} \\ \hline
                         JS                       & 89.23                          & 70.42                          & 78.72                           & 93.23                          & 93.62                          & 93.42                           & 97.18                          & 92.29                          & 94.67                           \\
Simple KL                         & 92.44                          & 98.82                          & 95.52                           & 92.20                          & 93.44                          & 92.82                           & 97.80                         & 96.71                          & 97.25                           \\ \hline 
DCdetector & 93.69 & 99.69 & \textbf{96.60} & 95.63 & 98.92 & \textbf{97.02} & 97.14 & 98.74 & \textbf{97.94} \\
\hline \hline
\end{tabular}}
\label{Tab: Ablation Metric Loss Function}
\end{table*}
\begin{table*}[h!]
\caption{Ablation studies on multi-scale patching results (window size=60). All results are in \%. The best ones are in \textbf{Bold}.}
{
\begin{tabular}{c|cccc|cccc|cccc|cc}
\hline \hline
\textbf{Dataset}                          & \multicolumn{4}{c|}{\textbf{MSL}}                     & \multicolumn{4}{c|}{\textbf{SMAP}}                    & \multicolumn{4}{c|}{\textbf{PSM}}                     & \multirow{2}{*}{\makecell[c]{\textbf{Mem} \\ \textbf{(GB)}}} & \multirow{2}{*}{\makecell[c]{\textbf{Time} \\ \textbf{(s)}}} \\ \cline{1-13}
\textbf{Metric} & \textbf{Acc} & \textbf{P} & \textbf{R} & \textbf{F1} & \textbf{Acc} & \textbf{P} & \textbf{R} & \textbf{F1} & \textbf{Acc} & \textbf{P} & \textbf{R} & \textbf{F1} &                                    &                                    \\ \hline
Patch Size = {[}1{]}                      & 97.98        & 92.77      & 88.29      & 90.48       & 98.96        & 93.91      & 98.31      & 96.06       & 98.82        & 97.27      & 98.07      & 97.66       & 16.9                               & 0.42                               \\
Patch Size = {[}3{]}                      & 98.64        & 92.39      & 95.34      & 93.84       & 98.59        & 94.65      & 94.43      & 94.54       & 97.22        & 96.95      & 91.84      & 94.33       & 6.0                                & 0.24                               \\
Patch Size = {[}5{]}                      & 98.91        & 92.55      & 97.87      & 95.14       & 98.92        & 94.40      & 97.44      & 95.90       & 98.75        & 97.22      & 97.84      & 97.53       & 3.2                                & 0.17                               \\
Patch Size = {[}1,3{]}                    & 98.30        & 93.19      & 90.77      & 91.96       & 98.98        & 94.42      & 97.87      & 96.11       & 98.83        & 96.96      & 98.42      & \textbf{97.68}       & 16.9                               & 0.59                               \\
Patch Size = {[}1,5{]}                    & 98.52        & 92.88      & 93.60      & 93.24       & 98.89        & 94.15      & 97.49      & 95.79       & 98.76        & 97.03      & 98.07      & 97.55       & 16.9                               & 0.46                               \\
Patch Size = {[}3,5{]}                    & 98.93        & 93.88      & 97.72      & \textbf{95.76}       & 98.89        & 94.61      & 96.95      & 95.77       & 98.38        & 97.00      & 96.54      & 96.77       & 6.0                                & 0.27                               \\
Patch Size = {[}1,3,5{]}                  & 98.44        & 91.52      & 93.78      & 92.64       & 99.03        & 93.72      & 99.10      & \textbf{96.34}       & 98.95        & 97.14      & 98.74      & 97.94       & 16.9                               & 0.71                              \\ \hline \hline 
\end{tabular}}
\label{Tab: Ablation multiscale results}
\end{table*}
\begin{table*}[h!]
\caption{Ablation studies on window size results (patch size=[3,5]). All results are in \%. The best ones are in \textbf{Bold}.}
{
\begin{tabular}{c|llll|llll|llll|cc}
\hline \hline
\textbf{Dataset}                        & \multicolumn{4}{c|}{\textbf{MSL}}                                                                                                     & \multicolumn{4}{c|}{\textbf{SMAP}}                                                                                                    & \multicolumn{4}{c|}{\textbf{PSM}}                                                                                                     & \multirow{2}{*}{\makecell[c]{\textbf{Mem} \\ \textbf{(GB)}}} & \multirow{2}{*}{\makecell[c]{\textbf{Time} \\ \textbf{(s)}}} \\ \cline{1-13}
\textbf{Metric} & \multicolumn{1}{c}{\textbf{Acc}} & \multicolumn{1}{c}{\textbf{P}} & \multicolumn{1}{c}{\textbf{R}} & \multicolumn{1}{c|}{\textbf{F1}} & \multicolumn{1}{c}{\textbf{Acc}} & \multicolumn{1}{c}{\textbf{P}} & \multicolumn{1}{c}{\textbf{R}} & \multicolumn{1}{c|}{\textbf{F1}} & \multicolumn{1}{c}{\textbf{Acc}} & \multicolumn{1}{c}{\textbf{P}} & \multicolumn{1}{c}{\textbf{R}} & \multicolumn{1}{c|}{\textbf{F1}} &                                    &                                    \\ \hline
Window size = 30                                      & 96.87                            & 92.39                          & 76.47                          & 83.68                           & 98.62                            & 93.93                          & 95.46                          & 94.69                           & 98.42                            & 97.42                          & 96.27                          & 96.84                           & 2.9                                & 0.17                               \\
Window size = 45                                      & 98.77                            & 92.80                          & 96.13                          & 94.44                           & 98.94                            & 94.24                          & 97.75                          & 95.96                           & 98.79                            & 97.01                          & 98.09                          & 97.55                           & 6.0                                & 0.22                               \\
Window size = 60                                      & 98.63                            & 91.82                          & 96.01                          & 93.87                           & 98.87                            & 94.87                          & 96.44                          & 95.65                           & 98.91                            & 97.04                          & 98.67                          & \textbf{97.85}                  & 6.1                                & 0.28                               \\
Window size = 75                                      & 98.79                            & 91.71                          & 97.93                          & 94.72                           & 98.97                            & 94.62                          & 97.60                          & 96.09                           & 98.79                            & 97.26                          & 98.20                          & 97.73                           & 7.6                                & 0.36                               \\
Window size = 90                                      & 98.94                            & 92.04                          & 98.82                          & 95.31                           & 98.99                            & 94.61                          & 97.69                          & 96.13                           & 98.74                            & 96.87                          & 98.05                          & 97.46                           & 7.6                                & 0.40                               \\
Window size = 105                                     & 99.06                            & 93.69                          & 99.69                          & \textbf{96.60}                  & 99.16                            & 94.69                          & 98.87                          & \textbf{96.74}                  & 98.57                            & 96.84                          & 97.37                          & 97.10                           & 18.5                               & 0.46                               \\
Window size = 120                                     & 98.95                            & 92.64                          & 98.74                          & 95.59                           & 99.08                            & 94.48                          & 98.48                          & 96.44                           & 98.80                            & 97.00                          & 98.24                          & 97.61                           & 18.5                               & 0.53                               \\
Window size = 135                                     & 98.44                            & 91.52                          & 94.45                          & 92.96                           & 99.09                            & 94.26                          & 98.91                          & 96.53                           & 98.70                            & 96.97                          & 98.17                          & 97.57                           & 24.4                               & 0.60                               \\
Window size = 150                                     & 98.49                            & 91.70                          & 95.02                          & 93.34                           & 98.93                            & 94.40                          & 97.48                          & 95.92                           & 98.55                            & 97.02                          & 97.18                          & 97.10                           & 24.4                               & 0.67                               \\
Window size = 165                                     & 98.64                            & 92.61                          & 95.68                          & 94.12                           & 99.01                            & 94.50                          & 98.07                          & 96.25                           & 98.77                            & 97.11                          & 98.05                          & 97.58                           & 24.4                               & 0.74                               \\
Window size = 180                                     & 98.68                            & 92.13                          & 96.13                          & 94.09                           & 98.99                            & 94.53                          & 97.73                          & 96.10                           & 98.67                            & 97.31                          & 97.87                          & 97.59                           & 24.4                               & 0.81                               \\
Window size = 195                                     & 98.50                            & 92.68                          & 94.18                          & 93.43                           & 98.95                            & 94.44                          & 97.60                          & 96.00                           & 98.66                            & 97.24                          & 97.88                          & 97.55                           & 24.5                               & 0.89                               \\
Window size = 210                                     & 98.03                            & 91.31                          & 90.88                          & 91.09                           & 98.58                            & 92.79                          & 96.29                          & 94.51                           & 98.39                            & 96.86                          & 96.59                          & 96.72                           & 24.5                               & 0.97  \\ \hline \hline                            
\end{tabular}}
\label{Tab: Ablation window size results}
\end{table*}
\begin{table*}[h!]
\caption{Ablation studies on attention head $H$ results (patch size=[3,5], window size=60). All results are in \%. The best ones are in \textbf{Bold}.}
{
\begin{tabular}{c|llll|llll|llll|cc}
\hline \hline
\textbf{Dataset}                          & \multicolumn{4}{c|}{\textbf{MSL}}                                                                                                     & \multicolumn{4}{c|}{\textbf{SMAP}}                                                                                                    & \multicolumn{4}{c|}{\textbf{PSM}}                                                                                                     & \multirow{2}{*}{\makecell[c]{\textbf{Mem} \\ \textbf{(GB)}}} & \multirow{2}{*}{\makecell[c]{\textbf{Time} \\ \textbf{(s)}}} \\ \cline{1-13}
\textbf{Metric} & \multicolumn{1}{c}{\textbf{Acc}} & \multicolumn{1}{c}{\textbf{P}} & \multicolumn{1}{c}{\textbf{R}} & \multicolumn{1}{c|}{\textbf{F1}} & \multicolumn{1}{c}{\textbf{Acc}} & \multicolumn{1}{c}{\textbf{P}} & \multicolumn{1}{c}{\textbf{R}} & \multicolumn{1}{c|}{\textbf{F1}} & \multicolumn{1}{c}{\textbf{Acc}} & \multicolumn{1}{c}{\textbf{P}} & \multicolumn{1}{c}{\textbf{R}} & \multicolumn{1}{c|}{\textbf{F1}} &                                    &                                    \\ \hline
$H$ = 1                                   & 98.63                            & 91.82                          & 96.01                          & \textbf{93.87}                  & 98.87                            & 94.87                          & 96.44                          & 95.65                           & 98.91                            & 97.04                          & 98.67                          & \textbf{97.85}                  & 6.1                                & 0.05                               \\
$H$ = 2                                   & 98.50                             & 92.13                          & 94.29                          & 93.20                            & 98.98                            & 93.99                          & 98.44                          & 96.16                           & 98.67                            & 97.16                          & 97.55                          & 97.36                           & 6.1                                & 0.17                               \\
$H$ = 4                                   & 98.67                            & 91.93                          & 95.71                          & 93.78                           & 98.97                            & 93.89                          & 98.46                          & 96.12                           & 98.69                            & 97.02                          & 97.79                          & 97.41                           & 9.8                                & 0.19                               \\
$H$ = 8                                   & 98.55                            & 91.30                           & 95.21                          & 93.21                           & 99.11                            & 94.87                          & 98.47                          & \textbf{96.63}                  & 98.63                            & 96.84                          & 97.73                          & 97.29                           & 9.8                                & 0.40    \\ \hline \hline                           
\end{tabular}}
\label{Tab: Ablation attention head number results}
\end{table*}
\begin{table*}[h!]
\caption{Ablation studies on embedding $d_{model}$ results (patch size=[3,5], window size=60). All results are in \%. The best ones are in \textbf{Bold}.}
{
\begin{tabular}{c|llll|llll|llll|cc}
\hline \hline
\textbf{Dataset}    & \multicolumn{4}{c|}{\textbf{MSL}}                                                                                                     & \multicolumn{4}{c|}{\textbf{SMAP}}                                                                                                    & \multicolumn{4}{c|}{\textbf{PSM}}                                                                                                     & \multirow{2}{*}{\makecell[c]{\textbf{Mem} \\ \textbf{(GB)}}} & \multirow{2}{*}{\makecell[c]{\textbf{Time} \\ \textbf{(s)}}} \\ \cline{1-13}
\textbf{Metric}     & \multicolumn{1}{c}{\textbf{Acc}} & \multicolumn{1}{c}{\textbf{P}} & \multicolumn{1}{c}{\textbf{R}} & \multicolumn{1}{c|}{\textbf{F1}} & \multicolumn{1}{c}{\textbf{Acc}} & \multicolumn{1}{c}{\textbf{P}} & \multicolumn{1}{c}{\textbf{R}} & \multicolumn{1}{c|}{\textbf{F1}} & \multicolumn{1}{c}{\textbf{Acc}} & \multicolumn{1}{c}{\textbf{P}} & \multicolumn{1}{c}{\textbf{R}} & \multicolumn{1}{c|}{\textbf{F1}} &                           &                           \\ \hline
$d_{model}$ = 128  & 98.47                            & 91.62                          & 94.59                          & 93.08                  & 99.10                             & 94.85                          & 98.34                          & \textbf{96.56}                  & 98.86                            & 97.13                          & 98.38                          & 97.75                           & 3.9                       & 0.05                      \\
$d_{model}$ = 256  & 98.79                            & 91.47                          & 98.02                          & \textbf{94.63}                  & 99.02                            & 94.25                          & 98.40                           & 96.28                           & 98.85                            & 96.98                          & 98.51                          & 97.74                           & 6.1                       & 0.10                       \\
$d_{model}$ = 512  & 98.63                            & 91.82                          & 96.01                          & 93.87                           & 98.87                            & 94.87                          & 96.44                          & 95.65                           & 98.91                            & 97.04                          & 98.67                          & 97.85                           & 10.3                      & 0.28                      \\
$d_{model}$ = 1024 & 98.13                            & 91.92                          & 90.81                          & 91.36                           & 98.97                            & 94.87                          & 97.27                          & 96.06                           & 98.97                            & 97.11                          & 98.83                          & \textbf{97.96}                  & 18.4                       & 0.92     \\ \hline \hline                
\end{tabular}}
\label{Tab: Ablation Embedding d_model results}
\end{table*}
\begin{table*}[h!]
\caption{Ablation studies on encoder layers $L$ results (patch size=[3,5], window size=60). All results are in \%. The best ones are in \textbf{Bold}.}
{
\begin{tabular}{c|llll|llll|llll|cc}
\hline \hline
\textbf{Dataset} & \multicolumn{4}{c|}{\textbf{MSL}}                                                                                                     & \multicolumn{4}{c|}{\textbf{SMAP}}                                                                                                    & \multicolumn{4}{c|}{\textbf{PSM}}                                                                                                     & \multirow{2}{*}{\makecell[c]{\textbf{Mem} \\ \textbf{(GB)}}} & \multirow{2}{*}{\makecell[c]{\textbf{Time} \\ \textbf{(s)}}} \\ \cline{1-13}
\textbf{Metric}  & \multicolumn{1}{c}{\textbf{Acc}} & \multicolumn{1}{c}{\textbf{P}} & \multicolumn{1}{c}{\textbf{R}} & \multicolumn{1}{c|}{\textbf{F1}} & \multicolumn{1}{c}{\textbf{Acc}} & \multicolumn{1}{c}{\textbf{P}} & \multicolumn{1}{c}{\textbf{R}} & \multicolumn{1}{c|}{\textbf{F1}} & \multicolumn{1}{c}{\textbf{Acc}} & \multicolumn{1}{c}{\textbf{P}} & \multicolumn{1}{c}{\textbf{R}} & \multicolumn{1}{c|}{\textbf{F1}} &                           &                           \\ \hline 
$L$ = 1              & 98.73                            & 91.76                          & 96.52                          & \textbf{94.08}                          & 98.94                            & 94.34                          & 97.69                          & 95.98                           & 96.92                            & 97.26                          & 90.33                          & 93.66                           & 6.0                         & 0.02                      \\
$L$ = 2              & 98.67                            & 98.67                          & 94.56                          & 93.72                           & 98.75                            & 93.93                          & 96.53                          & 95.22                           & 98.88                            & 97.24                          & 98.35                          & 97.79                           & 6.0                         & 0.04                      \\
$L$ = 3              & 98.63                            & 91.82                          & 96.01                          & 93.87                           & 98.87                            & 94.87                          & 96.44                          & 95.65                           & 98.91                            & 97.04                          & 98.67                          & \textbf{97.85}                  & 6.1                       & 0.10                       \\
$L$ = 4              & 98.33                            & 91.52                          & 92.72                          & 92.11                           & 99.01                            & 95.03                          & 97.42                          & 96.21                           & 98.88                            & 97.00                             & 98.53                          & 97.76                           & 6.1                       & 0.19                      \\
$L$ = 5              & 98.34                            & 91.38                          & 92.91                          & 92.14                           & 99.04                            & 94.23                          & 98.63                          & \textbf{96.38}                           & 98.83                            & 96.97                          & 98.41                          & 97.69                           & 6.1                       & 0.24        \\ \hline \hline             
\end{tabular}}
\label{Tab: Ablation Encoder layers results}
\end{table*}
\begin{table*}[]
\caption{Ablation studies on anomaly threshold $\delta$ results (patch size=[3,5], window size=60). All results are in \%. The best ones are in \textbf{Bold}.}
{
\begin{tabular}{c|llll|llll|llll}
\hline \hline
\textbf{Dataset} & \multicolumn{4}{c|}{\textbf{MSL}}                                                                                                     & \multicolumn{4}{c|}{\textbf{SMAP}}                                                                                                    & \multicolumn{4}{c}{\textbf{PSM}}                                                                                                     \\ \hline
\textbf{Metric}  & \multicolumn{1}{c}{\textbf{Acc}} & \multicolumn{1}{c}{\textbf{P}} & \multicolumn{1}{c}{\textbf{R}} & \multicolumn{1}{c|}{\textbf{F1}} & \multicolumn{1}{c}{\textbf{Acc}} & \multicolumn{1}{c}{\textbf{P}} & \multicolumn{1}{c}{\textbf{R}} & \multicolumn{1}{c|}{\textbf{F1}} & \multicolumn{1}{c}{\textbf{Acc}} & \multicolumn{1}{c}{\textbf{P}} & \multicolumn{1}{c}{\textbf{R}} & \multicolumn{1}{c}{\textbf{F1}} \\ \hline
$\delta$=0.5     & 96.62                            & 95.69                          & 72.23                          & 82.32                           & 98.85                            & 96.60                          & 94.40                          & 95.49                           & 98.45                            & 98.76                          & 95.03                          & 96.86                           \\
$\delta$=0.6     & 98.08                            & 94.75                          & 87.20                          & 90.80                           & 99.12                            & 96.66                          & 96.53                          & 96.59                           & 98.71                            & 98.27                          & 96.56                          & 97.41                           \\
$\delta$=0.7     & 98.80                            & 93.65                          & 95.47                          & 94.55                           & 99.29                            & 95.73                          & 98.88                          & \textbf{97.28}                  & 98.98                            & 98.14                          & 97.81                          & \textbf{97.97}                  \\
$\delta$=0.8     & 98.83                            & 93.12                          & 96.40                          & \textbf{94.73}                  & 99.06                            & 94.29                          & 98.72                          & 96.45                           & 98.88                            & 97.71                          & 97.85                          & 97.78                           \\
$\delta$=0.9     & 98.42                            & 91.90                          & 93.75                          & 92.82                           & 98.82                            & 93.81                          & 97.32                          & 95.54                           & 98.92                            & 97.42                          & 98.31                          & 97.86                           \\
$\delta$=1.0     & 98.33                            & 92.58                          & 92.04                          & 92.31                           & 98.90                            & 93.30                          & 98.56                          & 95.86                           & 98.91                            & 97.04                          & 98.67                          & 97.85                          \\ \hline \hline
\end{tabular}}
\label{Tab: Ablation Anomaly Threshold results}
\end{table*}

\end{document}